\documentclass[11pt]{article}
\usepackage{jmlr2e}
\usepackage{abstract}
\usepackage{array}
\usepackage{bbding}
\usepackage{multirow}
\usepackage{adjustbox}
\usepackage{bbm}
\usepackage{float}
\usepackage{amsfonts}
\usepackage{graphicx}
\graphicspath{{graphics/}}
\usepackage{subfigure}
\usepackage{float}

\usepackage{tabularx,booktabs}
\usepackage{multirow}
\newcolumntype{Y}{>{\centering\arraybackslash}X}

\usepackage[ruled,vlined,algo2e]{algorithm2e}
\usepackage{algorithm}
\usepackage{algorithmic}
\usepackage{natbib}

\usepackage{lscape}
\usepackage{comment}
\usepackage{bm}
\usepackage[title]{appendix}
\usepackage{longtable}
\usepackage{mathtools}
\usepackage{dsfont}
\usepackage{fancyhdr}

\usepackage{stackengine}
\usepackage{textcomp}
\usepackage{stfloats}
\usepackage{verbatim}
\usepackage{xcolor}
\definecolor{darkblue}{rgb}{0.0,0.5,0.5}
\definecolor{blue}{rgb}{0.0,0.0,1}

\usepackage[colorlinks]{hyperref}
\hypersetup{colorlinks,breaklinks,linkcolor=blue,urlcolor=blue,anchorcolor=blue,citecolor=blue}
\usepackage{cleveref}
\usepackage{booktabs,caption}
\usepackage{multirow}
\usepackage{lscape}
\usepackage{epstopdf}
\usepackage{lineno}
\usepackage{microtype}

\usepackage{graphicx}
\usepackage{subfigure}
\usepackage{booktabs} 
\usepackage{bbm}

\usepackage{amsfonts}
\usepackage{xurl}
\usepackage{stackengine}
\usepackage{tikz}
\usetikzlibrary{decorations.pathreplacing}
\usetikzlibrary{positioning,arrows.meta,quotes}
\usetikzlibrary{shapes,snakes}
\usetikzlibrary{bayesnet}
\tikzset{>=latex}
\tikzstyle{plate caption} = [caption, node distance=0, inner sep=0pt, below left=5pt and 0pt of #1.south]
\usepackage[normalem]{ulem}
\usepackage{multirow}

\newtheorem{assumption}{Assumption}

\newcommand{\vectorize}{\operatorname{vec}}
\usepackage{hyperref}
\usepackage{url}
\usepackage{enumitem}
\setlist{
  listparindent=\parindent,
  parsep=0pt,
}

\firstpageno{1}

\begin{document}

\title{Probabilistic Traffic Forecasting with Dynamic Regression}

\author{\name Vincent Zhihao Zheng \email zhihao.zheng@mail.mcgill.ca\\
       \addr Department of Civil Engineering\\
       McGill University\\
       817 Sherbrooke Street West, Montréal, Québec H3A 0C3, Canada
       \AND
       \name Seongjin Choi \email chois@umn.edu \\
       \addr Department of Civil, Environmental, and Geo- Engineering\\
       University of Minnesota\\
       500 Pillsbury Dr. SE, Minneapolis, MN 55455, USA
       \AND
       \name Lijun Sun\thanks{Corresponding author.} \email lijun.sun@mcgill.ca \\
       \addr Department of Civil Engineering\\
       McGill University\\
       817 Sherbrooke Street West, Montréal, Québec H3A 0C3, Canada}

\editor{}

\maketitle

\begin{abstract}
This paper proposes a dynamic regression (DR) framework that enhances existing deep spatiotemporal models by incorporating structured learning for the error process in traffic forecasting. The framework relaxes the assumption of time independence by modeling the error series of the base model (i.e., a well-established traffic forecasting model) using a matrix-variate autoregressive (AR) model. The AR model is integrated into training by redesigning the loss function. The newly designed loss function is based on the likelihood of a non-isotropic error term, enabling the model to generate probabilistic forecasts while preserving the original outputs of the base model. Importantly, the additional parameters introduced by the DR framework can be jointly optimized alongside the base model. Evaluation on state-of-the-art (SOTA) traffic forecasting models using speed and flow datasets demonstrates improved performance, with interpretable AR coefficients and spatiotemporal covariance matrices enhancing the understanding of the model.

\end{abstract}


\section{Introduction}
Traffic forecasting is a multivariate and multistep time series forecasting challenge that is a fundamental task within intelligent transportation systems (ITS). It has numerous applications, including trip planning, travel time estimation, and traffic flow management, among others \citep{vlahogianni2014short}. Imagine a traffic network equipped with $N$ sensors, collecting traffic data within a matrix of dimensions $N \times T$ over a total observation span of $T$. The ultimate objective of traffic forecasting is to anticipate the traffic conditions for $Q$ future intervals based on the most recent $P$ historical intervals.

\begin{figure}[htbp]
  \centering
  \includegraphics[width=0.75\textwidth, interpolate=false]{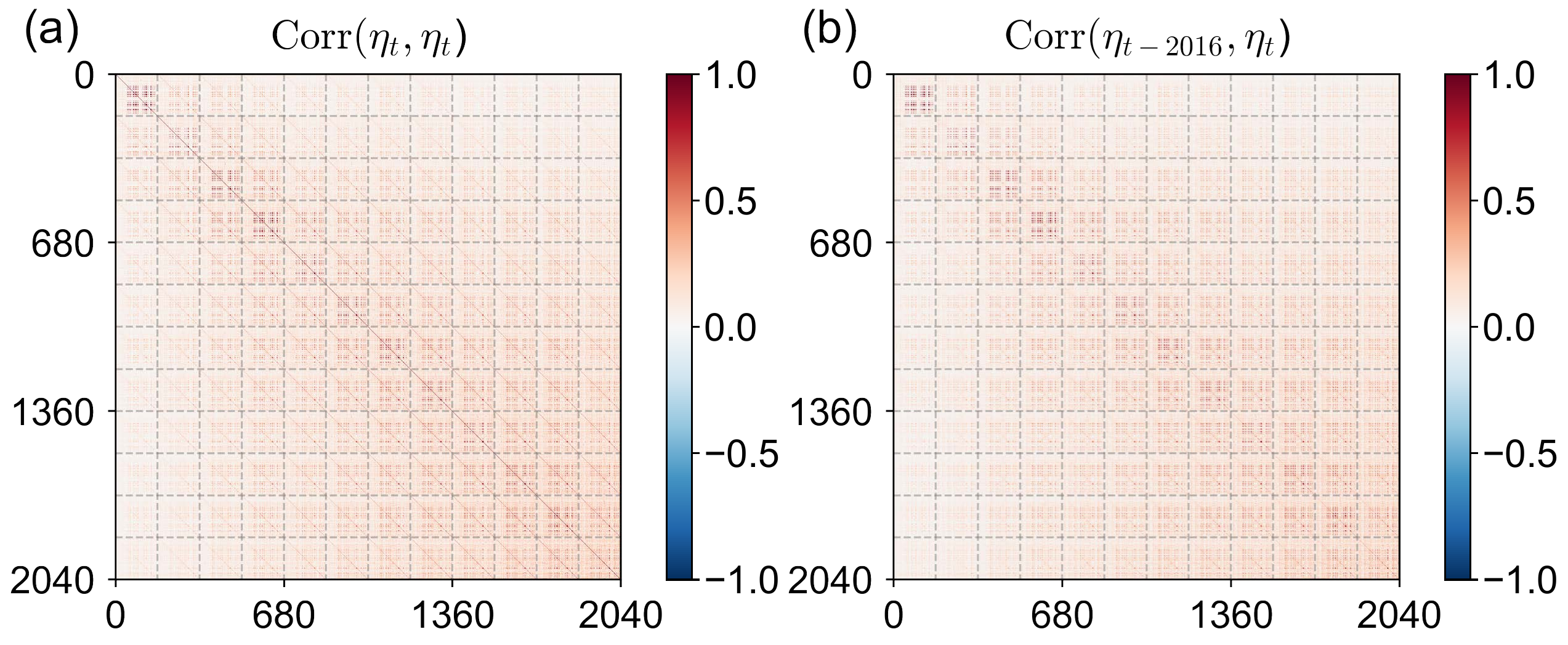}
  \caption{Correlation matrices of residuals obtained from the original ASTGCN model on the PEMS08 dataset, where $\boldsymbol{\eta}_t=\operatorname{vec}(\boldsymbol{R}_t) \in \mathbb{R}^{NQ}$ is the vectorized residual ($N=170$ and $Q=12$). (a) Contemporaneous correlations. (b) Cross-correlations at lag 2016 (i.e., one week).}
  \label{fig:fig1_stsgcn_pm8_res}
\end{figure}

Deep learning (DL) models are increasingly popular in traffic forecasting due to their ability to capture complex nonlinearity and spatiotemporal structures present in traffic data. Noteworthy deep spatiotemporal forecasting models, including STGCN \citep{yu2017spatio}, DCRNN \citep{li2017diffusion}, Graph Wavenet \citep{wu2019graph}, and STSGCN \citep{song2020spatial}, have demonstrated strong performance. DL models typically use mean absolute error (MAE) or mean squared error (MSE) for training, assuming that errors are temporally independent and follow an isotropic distribution. However, these assumptions do not align with real-world datasets. First, the omission of certain features often gives rise to time-correlated residuals. For example, in traffic speed forecasting, time-varying factors, such as weather conditions and vehicle flow rates, are often overlooked, leading to temporally correlated errors. Second, spatiotemporal forecasting suggests the potential existence of contemporaneous correlations within the errors, which contradicts the assumption of isotropic errors. For instance, it is intuitive that the variance of predicted values would increase from the 5-minute-ahead prediction to the 60-minute-ahead prediction. Neglecting these factors could detrimentally impact model performance.

As demonstrated in Figure~\ref{fig:fig1_stsgcn_pm8_res}, the contemporaneous correlations, $\operatorname{Corr}(\boldsymbol{\eta}_t,\boldsymbol{\eta}_t)$, and the lag-2016 cross-correlation, $\operatorname{Corr}(\boldsymbol{\eta}_{t-2016},\boldsymbol{\eta}_{t})$, calculated based on the residuals of ASTGCN \citep{song2020spatial} trained on the PEMS08 traffic flow dataset, exhibit distinct correlation structures. The strong off-diagonal elements in $\operatorname{Corr}(\boldsymbol{\eta}_t,\boldsymbol{\eta}_t)$ suggest that the errors are unlikely isotropic (Figure~\ref{fig:fig1_stsgcn_pm8_res} (a)). The cross-correlation $\operatorname{Corr}(\boldsymbol{\eta}_{t-2016},\boldsymbol{\eta}_{t})$ includes both the autocorrelation of errors $\operatorname{Corr}(\eta_{i,t-2016},\eta_{i,t})$ and the cross-lag correlation $\operatorname{Corr}(\eta_{i,t-2016},\eta_{j,t})$ between all pairs of components in the multivariate series. Figure~\ref{fig:fig1_stsgcn_pm8_res} (b) indicates significant cross-correlations in $\operatorname{Corr}(\boldsymbol{\eta}_{t-2016},\boldsymbol{\eta}_{t})$, likely resulting from omitted influential covariates, such as traffic congestion information, that significantly affect observed flow \citep{cheng2021incorporating}. While there are many modern techniques to identify and incorporate these covariates into deep learning models, our method offers a model-agnostic alternative. By leveraging the statistical attributes of the error process, our approach enhances model performance without relying on extensive features (some of which may be impractical to obtain) or perfectly tuned deep learning models.

This study presents an efficient method for adjusting and modeling correlated errors using a dynamic regression framework. The method is designed for seamless integration into any deep learning model applied to spatiotemporal traffic forecasting (i.e., multivariate sequence-to-sequence (Seq2Seq) forecasting). The approach assumes that the error of the base model follows a matrix-variate AR process, which accounts for the error cross-correlation problem and can be easily integrated into the model training process. In addition to gaining insights into the coefficients of the AR process, the method learns a non-isotropic covariance matrix effectively for the error term, which is assumed to follow a multivariate normal distribution. Notably, the parameters of the error AR module and the error covariance matrix can be jointly optimized with the parameters of the base model. The primary contributions of this study are outlined as follows:
\begin{itemize}
\item We propose utilizing a bi-linear AR structure for the matrix-valued error process to effectively address cross-correlation. By incorporating seasonal lags into the AR structure, we capture the characteristic seasonal patterns present in traffic data, thereby improving model accuracy and reliability.
\item We model the error term in the DR framework using a multivariate normal distribution with a non-isotropic covariance matrix. The negative log-likelihood of this distribution serves as the loss function for model optimization. This approach yields an interpretable covariance matrix, which can be used for probabilistic forecasting with uncertainty quantification in the Seq2Seq problem.
\item The proposed method is model-agnostic, enhancing existing DL models by leveraging the statistical attributes of the error process, which is applicable to any spatiotemporal traffic forecasting problem. We evaluate our method using several SOTA deep traffic forecasting models and consistently observe improvements across various metrics.
\end{itemize}

The subsequent sections of this paper are structured as follows. In Section \ref{sec:lr_reviews}, we conduct a comprehensive review of related works, covering deep traffic forecasting models, probabilistic forecasting methods, and the challenges associated with error correlations. Section \ref{sec:method} provides a detailed account of the traffic forecasting problem and introduces our proposed dynamic regression framework. In Section \ref{sec:exp}, we present the experimental setup and report the obtained results. Section \ref{sec:discuss} discusses some conceptual questions related to the proposed method. Section \ref{sec:summary} summarizes key findings and concludes the paper.

\section{Related Works}\label{sec:lr_reviews}
\subsection{Deep Learning for Traffic Forecasting}

In this section, we review key deep learning frameworks for traffic forecasting, a spatiotemporal forecasting (STF) problem. Traffic data is characterized by complex spatial dependencies, such as interactions between road segments or sensors, and temporal patterns, including daily or weekly cycles. To effectively model these intertwined dependencies, most existing works leverage Graph Neural Networks (GNNs), which excel at capturing the spatial relationships in traffic networks. These models are often combined with RNNs or Temporal Convolutional Networks (TCNs) to model temporal dynamics.

Graph-based approaches for traffic forecasting can be broadly categorized into spectral graph convolution and diffusion graph convolution methods \citep{ye2020build}. Spectral graph convolution operates in the frequency domain by leveraging the graph Laplacian to capture spatial dependencies through spectral decomposition. For instance, STGCN \citep{yu2017spatio} combines Graph Convolutional Networks (GCNs) to model spatial correlations with Convolutional Neural Networks (CNNs) to capture temporal dependencies. GCNs are particularly effective for incorporating graph structures into spatial modeling, while CNNs enable faster training through parallel computation, avoiding the sequential nature of RNNs. Building on STGCN, ASTGCN \citep{guo2019attention} introduces a spatiotemporal attention mechanism to preprocess traffic data, enabling the model to better identify and focus on relevant temporal and spatial features before passing the data to the convolutional layers.

In contrast, diffusion graph convolution operates directly in the vertex domain, modeling spatial dependencies through localized diffusion processes. DCRNN \citep{li2017diffusion} exemplifies this approach by integrating a diffusion convolution operation to model spatial dependencies, combined with Gated Recurrent Units (GRUs) for temporal modeling. Some models, such as STGCN and ASTGCN, rely on a pre-defined adjacency matrix that remain fixed during training, limiting flexibility. To address this, Graph WaveNet \citep{wu2019graph} employs an adaptive adjacency matrix that learns the graph structure by treating the matrix entries as trainable parameters. This model integrates dilated causal convolution as Temporal Convolutional Networks (TCNs) for temporal dependency modeling and GCNs for spatial dependency. Earlier GCN-based models treated spatial and temporal dependencies separately, limiting their ability to capture spatiotemporal interactions. To overcome this limitation, STSGCN \citep{song2020spatial} connects individual spatial graphs of adjacent time steps into a unified graph, enabling simultaneous learning of spatial and temporal information. This approach has demonstrated superior performance compared to previous GCN-based frameworks by fully leveraging spatiotemporal interactions. In D\textsuperscript{2}STGNN \citep{shao2022decoupled}, the authors decompose traffic signals into diffusion signals and inherent signals, processed separately using specialized modules. The diffusion module utilizes spatial-temporal localized convolutional layers to capture dynamic spatial dependencies, while the inherent module incorporates GRUs and self-attention layers to model temporal patterns intrinsic to the data. Additionally, the framework constructs adjacency matrices at different time steps to effectively capture the dynamic correlations among nodes.

Other SOTA models include GMAN \citep{zheng2020gman}, AGCRN \citep{bai2020adaptive}, and ASTGNN \citep{guo2021learning}, among others. For a comprehensive overview of STF models, we refer readers to the recent survey by \citet{shao2024exploring}.

\subsection{Probabilistic Forecasting Methods}

Probabilistic forecasting is central to time series analysis. One effective approach is to design models that output a set of quantiles rather than a single point forecast. For instance, \citet{wen2017multi} introduced MQ-RNN, a Seq2Seq RNN framework that directly forecasts quantiles. The model parameters are optimized using the quantile loss function. Similarly, \citet{toubeau2018deep} applied this methodology for probabilistic forecasting in power markets.

An alternative approach to probabilistic forecasting involves estimating the probability density functions (PDF) of target variables. This method typically assumes a simple distribution form for the target variables and uses neural networks to output the distribution parameters. A prominent example is DeepAR \citep{salinas2020deepar}, which employs RNNs to model the transition of hidden states. These hidden states condition the generation of parameters for a Gaussian distribution, from which prediction samples are drawn. Similarly, GPVar \citep{salinas2019high} uses a multivariate Gaussian distribution to model the joint distribution of multivariate time series. The authors utilize a Gaussian copula to transform the original time series observations into Gaussian random variables, with RNNs parameterizing the output distribution.

Neural networks can also be designed to learn probabilistic model parameters rather than assuming a fixed distribution. \citet{rangapuram2018deep} proposed a deep state space model (SSM), utilizing RNNs to output SSM parameters, which allows for drawing prediction samples from the observation model of the SSM. Similarly, \citet{de2020normalizing} introduced the Normalizing Kalman Filter (NKF) model, enhancing the classical linear Gaussian state space model (LGM) with normalizing flows (NFs). These NFs model non-linear dynamics on top of the observation model, allowing PDF estimation of observations. The parameters of the NKF model are generated by RNNs. Additionally, \citet{wang2019deep} proposed a probabilistic forecasting model based on deep factors. This model incorporates both a fixed global component and a random component, with the latter encompassing classical probabilistic time series models. The combination of these components forms the latent function, which conditions a parametric distribution for generating observations.

Some approaches focus on enhancing expressive conditioning for probabilistic forecasting. One such method involves using Transformers instead of RNNs to model latent dynamics, thereby relaxing the Markovian assumption in RNNs \citep{tang2021probabilistic}. Another avenue for improvement involves adopting more flexible distribution forms, such as normalizing flows \citep{rasul2020multivariate}, diffusion models \citep{rasul2021autoregressive}, and Copula-based models \citep{drouin2022tactis, ashok2023tactis}.

Our approach stands out from previously mentioned probabilistic models in two key aspects. Firstly, it extends probabilistic forecasting to multivariate Seq2Seq traffic forecasting, where the target outcome is a matrix rather than a scalar or vector. Secondly, it can seamlessly integrate with existing deterministic traffic forecasting models without altering their outputs.

\subsection{Adjusting for Correlated Errors}

Correlated errors in time series data have been extensively studied in econometrics through models with precise forms \citep{durbin1950testing,ljung1978measure,breusch1978testing,godfrey1978testing,cochrane1949application,prais1954trend,beach1978maximum,prado2021time,hyndman2018forecasting,hamilton2020time}. Recent advances in DL models have increased interest in learning and adapting the error process. Two primary statistical approaches are commonly employed for modeling the error process: (i) capturing autocorrelation (or cross-correlation in multivariate data), and (ii) learning contemporaneous correlation in errors.

To address autocorrelation or cross-correlation, \citet{sun2021adjusting} proposed a reparametrization strategy for the input and output of a neural network used in time series forecasting. This reparametrization inherently employs a first-order error AR process through a linear regression framework. The method effectively enhances performance for various DL-based one-step-ahead forecasting models across a diverse range of time series datasets, enabling joint parameter optimization of both base and error regressors. Additionally, \citet{kim2022residual} introduced a lightweight DL module designed to calibrate cross-correlation within predictions from pre-trained traffic forecasting models. This calibration module uses recent observed residuals and predictions to anticipate future residuals, ultimately enhancing the performance of numerous traffic forecasting models, especially on traffic speed datasets. \citet{zheng2024better} introduced a probabilistic forecasting model trained using a Generalized Least Squares (GLS) loss, which explicitly accounts for the time-varying autocorrelation of batched error terms. This approach was later extended to the multivariate scenario, accounting for error cross-correlation through the use of an efficient parameterization of the covariance matrix \citep{zheng2024multivariate}.

Regarding learning contemporaneous correlation, \citet{jia2020residual} emphasized that assuming independence in the residuals of a node regression problem is unwarranted. They advocated for modeling contemporaneous correlation using a multivariate Gaussian distribution. This method, termed residual propagation in GNNs, adjusts predictions of unknown nodes based on known node labels. Similarly, \citet{huang2020combining} introduced a correct-and-smooth approach, serving as a post-processing scheme to rectify correlated residuals in GNNs, focusing on a classification task. Additionally, \citet{choi2022spatiotemporal} proposed a dynamic mixture of matrix normal Gaussian as a regularization technique to address contemporaneous correlation within the errors of Seq2Seq models.

Applying existing methods directly to train a multivariate Seq2Seq model is challenging. A key challenge is that the error term becomes an \(N \times Q\) matrix-variate time series. The cohesive learning of this process and the base model constitutes a primary challenge. Our work differs from \citet{sun2021adjusting} and \citet{kim2022residual} in several important ways. Firstly, we extend the scope of \citet{sun2021adjusting} to encompass multivariate Seq2Seq forecasting, while avoiding the reparametrization of input and output. Secondly, our approach allows for the concurrent optimization of both the DR parameters and the parameters of the base model, unlike the post-hoc adjustment in \citet{kim2022residual}. Lastly, our method is constructed based on a probabilistic formulation with the capability for uncertainty quantification, setting our work apart from the aforementioned studies. The works of \citet{zheng2024better} and \citet{zheng2024multivariate} model the correlated errors in probabilistic forecasting but do not directly apply to the Seq2Seq task.

\section{Methodology}\label{sec:method}
\begin{figure}[t]
  \centering
  \includegraphics[width=0.75\textwidth, interpolate=false]{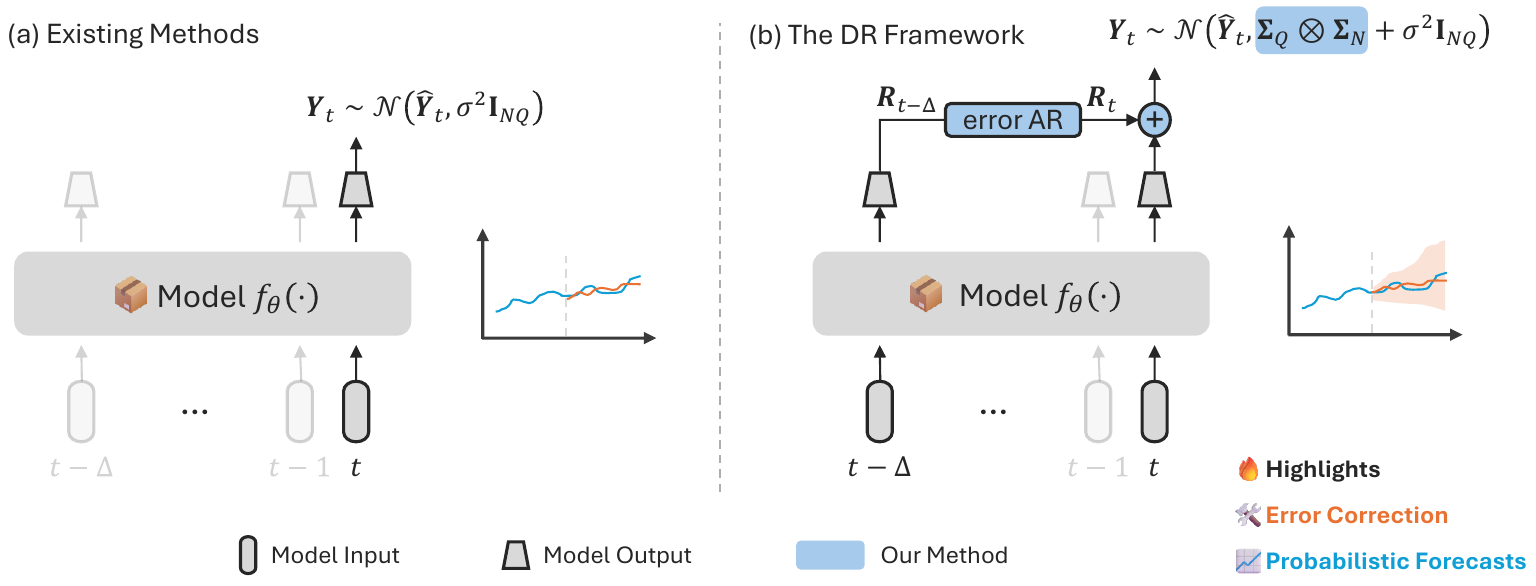}
  \caption{Overview of the proposed DR framework, highlighting its structure and comparison with conventional methods.}
  \label{fig:flow_chart}
\end{figure}

This section formulates the multivariate Seq2Seq traffic forecasting problem and outlines the dynamic regression framework for adjusting error cross-correlation and capturing contemporaneous correlations. Figure~\ref{fig:flow_chart} presents the overview of the proposed DR framework.

\subsection{Traffic Forecasting}

A traffic network can be defined as a directed graph $\mathcal{G}=\left(\mathcal{V}, \mathcal{E}, \mathbf{W}\right)$, where $\mathcal{V}$ with $\vert\mathcal{V}\vert=N$ is a set of nodes representing traffic sensors; $\mathcal{E}$ is a set of links connecting these nodes; $\mathbf{W} \in \mathbb{R}^{N \times N}$ is a weighted adjacency matrix characterizing the proximity of nodes. Let $\boldsymbol{z}_{t} \in \mathbb{R}^{N}$ denote the vector of observed traffic states at time $t$. The traffic forecasting problem involves learning a function that maps data from a context horizon of $P$ past time steps to a prediction horizon of $Q$ future time steps.

Let $\boldsymbol{X}_{t}=\left[\boldsymbol{z}_{t-P+1}, \dots, \boldsymbol{z}_{t}\right] \in \mathbb{R}^{N \times P}$ and $\boldsymbol{Y}_{t}=\left[\boldsymbol{z}_{t+1}, \dots, \boldsymbol{z}_{t+Q}\right] \in \mathbb{R}^{N \times Q}$, we have
\begin{equation}
\label{eqn:forecsating_mean}
    {\boldsymbol{Y}}_{t}=f_\theta(\boldsymbol{X}_{t}, \mathbf{W})+\boldsymbol{R}_t,
\end{equation}
where $f_\theta(\boldsymbol{X}_{t}, \mathbf{W})$ is a DL model that generates the predicted mean and $\boldsymbol{R}_t\in \mathbb{R}^{N\times Q}$ is the error term. For simplicity, we will omit $\mathbf{W}$ for the remainder of the paper. In many cases, $f_\theta(\cdot)$ is trained with MAE/MSE as the loss function to obtain an optimal parameter set \(\theta^\star\), e.g., a MSE loss can be written as:
\begin{equation}
\mathcal{L}=\sum_{t} \|\boldsymbol{R}_t\|_F^2 = \sum_{t} \|\boldsymbol{Y}_t-f_\theta(\boldsymbol{X}_{t})\|_F^2.
\end{equation}
This loss function simply assumes: (i) $\boldsymbol{R}_t$ is temporally independent, i.e., there is no correlation between $\boldsymbol{R}_s$ and $\boldsymbol{R}_t$ when $s\neq t$; and (ii) entries in $\boldsymbol{R}_t$ follow a zero-mean isotropic Gaussian distribution with no concurrent correlations, i.e.,  $\boldsymbol{\eta}_t=\operatorname{vec}(\boldsymbol{R}_t) \sim \mathcal{N}(\mathbf{0}, \sigma^2\mathbf{I}_{NQ})$. Likewise, using MAE as the loss function corresponds to assuming entries in $\boldsymbol{R}_t$ follow a Laplacian distribution.

\subsection{Modeling Error with Matrix-valued Autoregression} \label{sec:dr}

Following the idea of dynamic regression \citep{hyndman2018forecasting}, we assume that the relationship between the input $\boldsymbol{X}_{t}$ and the output $\boldsymbol{Y}_{t}$ is not fully captured by $f_\theta(\cdot)$, and the error $\boldsymbol{R}_t$ is governed by a temporal process. For example, for a one-step-ahead prediction model (i.e., $\boldsymbol{\eta}_t=\boldsymbol{R}_t$ as $Q=1$), it is straightforward to model $\boldsymbol{\eta}_t$ using a $p$-th order vector autoregressive model:
\begin{equation}\label{equ:var}
\boldsymbol{\eta}_t = \boldsymbol{C}_1 \boldsymbol{\eta}_{t-1}+\ldots +\boldsymbol{C}_p \boldsymbol{\eta}_{t-p} + \boldsymbol{\epsilon}_t,
\end{equation}
where $\boldsymbol{C}_l\in \mathbb{R}^{N\times N}$ ($l=1,\ldots,p$) are the regression coefficient, and $\boldsymbol{\epsilon}_t$ is a white-noise process.

However, for a Seq2Seq model with $Q>1$, $\boldsymbol{R}_t$ cannot be directly modeled using Eq.~\eqref{equ:var}, as $\{\boldsymbol{R}_{t-l}\,|\,l=1,\ldots, Q-1\}$ will not be entirely accessible due to overlapping. To address this issue, we model the relation between $\boldsymbol{R}_t$ and its accessible lagged values, i.e., $\boldsymbol{R}_{t-\Delta _l}$ for $l=1,\ldots,p$ as long as $\Delta _l\ge Q$. Therefore, we have the model for the vector $\boldsymbol{\eta}_{t}$ as:
\begin{equation}\label{equ:season}
\boldsymbol{\eta}_t = \boldsymbol{C}_1 \boldsymbol{\eta}_{t-\Delta _1}+\ldots+ \boldsymbol{C}_p \boldsymbol{\eta}_{t-\Delta _p} + \boldsymbol{\epsilon}_t,
\end{equation}
where $\boldsymbol{C}_l\in \mathbb{R}^{NQ\times NQ}$ and the white-noise covariance matrix is also of size $NQ\times NQ$. As traffic data often exhibit strong daily and weekly patterns, for simplicity we only introduce a single lagged residual component $\boldsymbol{R}_{t-\Delta }$, where $\Delta $ is a predetermined lag (e.g., one day or one week) exhibiting strong correlations with the present time.
\begin{assumption}\label{asmp1}
    The error \(\boldsymbol{R}_{t}\) of the base deep learning model follows a first-order bi-linear autoregressive process (Eq.~\eqref{eqn:matrix_ar1}).
\end{assumption}
A notable limitation of the aforementioned formulation is the introduction of an excessive number of parameters within $\boldsymbol{C}$. For improved scalability of this approach, we assume $\boldsymbol{R}_{t}$ follows a first-order bi-linear AR model \citep{chen2021autoregressive,hsu2021matrix}:
\begin{equation}
\label{eqn:matrix_ar1}
    \boldsymbol{R}_t=\boldsymbol{A} \boldsymbol{R}_{t-\Delta } \boldsymbol{B}+\boldsymbol{E}_t,
\end{equation}
where $\boldsymbol{A}\in\mathbb{R}^{N\times N}$ and $\boldsymbol{B}\in\mathbb{R}^{Q\times Q}$ are regression coefficients. With Assumption \ref{asmp1}, there are two error terms: \(\boldsymbol{R}_t\), the error from the base DL model, and \(\boldsymbol{E}_t\), the error from the AR model. In the DR framework, only the AR model error \(\boldsymbol{E}_t\) is assumed to be time-independent. The vectorized version of Eq.~\eqref{eqn:matrix_ar1} becomes
\begin{equation}
\label{eqn:kron_ar1}
    \boldsymbol{\eta}_t=(\boldsymbol{B}^\top \otimes \boldsymbol{A}) \boldsymbol{\eta}_{t-\Delta } +\boldsymbol{\epsilon}_t,
\end{equation}
and we can see that the bi-linear formulation is equivalent to imposing a Kronecker product structure on the coefficient matrix $\boldsymbol{C}=\boldsymbol{B}^\top\otimes\boldsymbol{A}$ in Eq.~\eqref{equ:season}, leading to a significant reduction in parameters. Combining Eqs.~\eqref{eqn:forecsating_mean} and \eqref{eqn:matrix_ar1}, we can obtain a new model that seamlessly integrates with the autocorrelated errors:
\begin{equation}
\begin{aligned}
\label{eqn:dr_model}
        \boldsymbol{Y}_t & = f_\theta(\boldsymbol{X}_{t}) + \boldsymbol{A} \boldsymbol{R}_{t-\Delta } \boldsymbol{B}+\boldsymbol{E}_t \\
        & = f_\theta(\boldsymbol{X}_{t}) + \boldsymbol{A}\left(\boldsymbol{Y}_{t-\Delta }-f_\theta(\boldsymbol{X}_{t-\Delta })\right)\boldsymbol{B} + \boldsymbol{E}_t,
\end{aligned}
\end{equation}
where $\boldsymbol{A}$ and $\boldsymbol{B}$, as trainable parameters, can be jointly updated with the base model $f_{\theta}(\cdot)$. When $\boldsymbol{R}_{t}$ and $\boldsymbol{R}_{t-\Delta}$ are uncorrelated (e.g., when $\boldsymbol{A}=\boldsymbol{0}$), The model reduces to its original implementation. If both $\boldsymbol{A}$ and $\boldsymbol{B}$ are identity matrices, the model implies a seasonal random walk for the error process. To enforce sparsity in $\boldsymbol{A}$ and $\boldsymbol{B}$, we introduce an $\ell_1$ regularization term into the loss function:
\begin{equation}
\label{eqn:loss}
\mathcal{L}_{\text{reg}}=
\frac{1}{N^2} \|\boldsymbol{A}\|_1+\frac{1}{Q^2} \|\boldsymbol{B}\|_1.
\end{equation}
Once $f_{\theta}(\cdot)$ and coefficients $\boldsymbol{A}$ and $\boldsymbol{B}$ are learned, prediction at time $t$ can be made by:
\begin{equation}
\label{eqn:mean_pred}
\hat{\boldsymbol{Y}}_t = f_\theta(\boldsymbol{X}_{t})+\boldsymbol{A}\left(\boldsymbol{Y}_{t-\Delta }-f_\theta(\boldsymbol{X}_{t-\Delta })\right)\boldsymbol{B}.
\end{equation}

In this paper, we refer to ``residuals'' as the difference between the observed values and the values predicted by the model, i.e., \(\boldsymbol{Y}_t - \hat{\boldsymbol{Y}}_t\). Here, \(\hat{\boldsymbol{Y}}_t\) is generated by a trained model \(g_{\Theta^\star}(\cdot)\). On the other hand, we refer to ``errors'' as the deviation of the observed values from the true underlying values, i.e., \(\boldsymbol{Y}_t - g_{\Theta}(\cdot)\), where \(g_{\Theta}(\cdot)\) is the true underlying model. Therefore, ``errors'' are theoretical quantities because the true underlying model is usually unknown in real-world applications. ``Residuals'', however, are observable quantities because they are based on the actual data and the fitted model.

\subsection{Spatiotemporal Covariance Structure for the Matrix Error Term}\label{sec:cov}

Following \cite{stegle2011efficient} and \cite{luttinen2012efficient}, we assume that the error $\boldsymbol{\epsilon}_t$ in Eq.~\eqref{eqn:kron_ar1} follows a zero-mean multivariate Gaussian distribution $\boldsymbol{\epsilon}_t=\vectorize{\left(\boldsymbol{E}_t\right)} \sim \mathcal{N}(\boldsymbol{0}_{NQ},\boldsymbol{\Sigma})$, where $\boldsymbol{\Sigma}=\boldsymbol{\Sigma}_Q \otimes \boldsymbol{\Sigma}_N + \sigma^2\mathbf{I}_{NQ}$ is a non-isotropic covariance matrix (Assumption \ref{asmp2}). Here, $\boldsymbol{\Sigma}_N$ and $\boldsymbol{\Sigma}_Q$ represent the contemporaneous covariance among the rows and columns of $\boldsymbol{E}_t$. The Kronecker structure significantly reduces the number of parameters in $\boldsymbol{\Sigma}$. The negative log-likelihood of this distribution is used as the new loss function:
\begin{equation}
\label{eq:nll_mnd}
\begin{aligned}
\mathcal{L}_{\text{nll}} & = - \ln p(\boldsymbol{E}_t\,|\,\boldsymbol{\Sigma}_Q,\boldsymbol{\Sigma}_N, \sigma^2) \\
& = \frac{1}{2} \left[\ln |\boldsymbol{\Sigma}| + \boldsymbol{\epsilon}_t^\top \boldsymbol{\Sigma}^{-1} \boldsymbol{\epsilon}_t + NQ \ln (2\pi)\right].
\end{aligned}
\end{equation}
\begin{assumption}\label{asmp2}
    The error \(\boldsymbol{\epsilon}_t=\vectorize{\left(\boldsymbol{E}_t\right)}\) in the dynamic regression framework follows a zero-mean multivariate Gaussian distribution with a covariance matrix \(\boldsymbol{\Sigma}\) parameterized as a Kronecker product plus a diagonal \(\boldsymbol{\Sigma}_Q \otimes \boldsymbol{\Sigma}_N + \sigma^2\mathbf{I}_{NQ}\) (Eq.~\eqref{eqn:svd}).
\end{assumption}
The large size (\(NQ \times NQ\)) of \(\boldsymbol{\Sigma}\) makes computing its inverse and determinant increasingly challenging as (\(N\)) grows. To efficiently optimize the log-likelihood, we apply the Sherman–Morrison–Woodbury identity and the matrix determinant lemma. This is facilitated by leveraging the low-rank decomposition of the covariance matrices \(\boldsymbol{\Sigma}_N = \boldsymbol{L}_N\boldsymbol{L}_N^\top\) and \(\boldsymbol{\Sigma}_Q = \boldsymbol{L}_Q\boldsymbol{L}_Q^\top\), where \(\boldsymbol{L}_N \in \mathbb{R}^{N \times R_n}\) and \(\boldsymbol{L}_Q \in \mathbb{R}^{Q \times R_q}\), with \(R_n\) and \(R_q\) being the rank parameters. Consequently,
\begin{equation}\label{eqn:svd}
\begin{aligned}
\boldsymbol{\Sigma}&=\boldsymbol{\Sigma}_Q \otimes \boldsymbol{\Sigma}_N + \sigma^2\mathbf{I}_{NQ}\\
&=\left(\boldsymbol{L}_Q\boldsymbol{L}_Q^\top\right) \otimes \left(\boldsymbol{L}_N\boldsymbol{L}_N^\top\right) + \sigma^2\mathbf{I}_{NQ}\\
&=\left(\boldsymbol{L}_Q \otimes \boldsymbol{L}_N \right)\left(\boldsymbol{L}_Q \otimes \boldsymbol{L}_N \right)^\top + \sigma^2\mathbf{I}_{NQ}\\
&=\boldsymbol{S}\boldsymbol{S}^\top+\boldsymbol{D},
\end{aligned}
\end{equation}
where \(\boldsymbol{S}=\boldsymbol{L}_Q \otimes \boldsymbol{L}_N\) and \(\boldsymbol{D}=\sigma^2\mathbf{I}_{NQ}\). The matrix inversion and determinant can be efficiently computed as
\begin{equation}\label{eqn:sigma_inv}
\boldsymbol{\Sigma}^{-1}=\left(\boldsymbol{D}+\boldsymbol{S}\boldsymbol{S}^\top\right)^{-1}
    =\boldsymbol{D}^{-1}-\boldsymbol{D}^{-1}\boldsymbol{S}\left(\mathbf{I}_{R}+\boldsymbol{S}^\top\boldsymbol{D}^{-1}\boldsymbol{S}\right)^{-1}\boldsymbol{S}^\top\boldsymbol{D}^{-1},
\end{equation}
\begin{equation}\label{eqn:sigma_det}
|\boldsymbol{\Sigma}|=\det{\left(\boldsymbol{D}+\boldsymbol{S}\boldsymbol{S}^\top\right)}
=\det{\left(\mathbf{I}_{R}+\boldsymbol{S}^\top\boldsymbol{D}^{-1}\boldsymbol{S}\right)}\det{\left(\boldsymbol{D}\right)},
\end{equation}
where \(R=R_nR_q\). Eq.~\eqref{eqn:sigma_inv} and Eq.~\eqref{eqn:sigma_det} simplify the computation of the inverse and determinant of $\boldsymbol{\Sigma}$ to calculating the inverse and determinant of an \(R \times R\) capacitance matrix $\left(\mathbf{I}_{R}^{-1}+\boldsymbol{S}^\top\boldsymbol{D}^{-1}\boldsymbol{S}\right)$. This can be efficiently achieved using its Cholesky factor \(\boldsymbol{L}_{C}\), where $\left(\mathbf{I}_{R}^{-1}+\boldsymbol{S}^\top\boldsymbol{D}^{-1}\boldsymbol{S}\right)=\boldsymbol{L}_{C}\boldsymbol{L}_{C}^\top$, and \(\boldsymbol{L}_{C}\) is a lower triangular matrix with positive diagonal entries. Specifically, the Mahalanobis term in Eq.~\eqref{eq:nll_mnd} can be simplified as
\begin{equation}\label{eqn:mahalanobis}
\begin{aligned}
    &\boldsymbol{\epsilon}_t^\top\boldsymbol{\Sigma}^{-1}\boldsymbol{\epsilon}_t\\
    =&\boldsymbol{\epsilon}_t^\top\boldsymbol{D}^{-1}\boldsymbol{\epsilon}_t-\boldsymbol{\epsilon}_t^\top\boldsymbol{D}^{-1}\boldsymbol{S}\left(\boldsymbol{L}_{C}\boldsymbol{L}_{C}^\top\right)^{-1}\boldsymbol{S}^\top\boldsymbol{D}^{-1}\boldsymbol{\epsilon}_t\\
    =&\boldsymbol{\epsilon}_t^\top\boldsymbol{D}^{-1}\boldsymbol{\epsilon}_t-\left(\boldsymbol{L}_{C}^{-1}\boldsymbol{S}^\top\boldsymbol{D}^{-1}\boldsymbol{\epsilon}_t\right)^\top\left(\boldsymbol{L}_{C}^{-1}\boldsymbol{S}^\top\boldsymbol{D}^{-1}\boldsymbol{\epsilon}_t\right)\\
    =&\boldsymbol{\epsilon}_t^\top\boldsymbol{D}^{-1}\boldsymbol{\epsilon}_t-\boldsymbol{k}_t^\top\boldsymbol{k}_t,
\end{aligned}
\end{equation}
where \(\boldsymbol{k}_t=\boldsymbol{L}_{C}^{-1}\boldsymbol{S}^\top\boldsymbol{D}^{-1}\boldsymbol{\epsilon}_t\). The computation of \(\boldsymbol{k}_t\) can be efficiently performed by solving a triangular system of linear equations, \(\boldsymbol{L}_{C}\boldsymbol{k}_t=\boldsymbol{S}^\top\boldsymbol{D}^{-1}\boldsymbol{\epsilon}_t\). Since \(\boldsymbol{D}\) is a diagonal, this method avoids direct matrix inversion. Additionally, the calculation of the log-determinant is greatly simplified as follows:
\begin{equation}\label{eqn:log_det}
\begin{aligned}
    \ln\lvert\boldsymbol{\Sigma}\rvert
    &=\ln\lvert\boldsymbol{D}+\boldsymbol{S}\boldsymbol{S}^\top\rvert\\
    &=\ln\lvert\mathbf{I}_{R}+\boldsymbol{S}^\top\boldsymbol{D}^{-1}\boldsymbol{S}\rvert + \ln\lvert\boldsymbol{D}\rvert\\
    &=2\sum^{R}_i\ln\left[\boldsymbol{L}_{C}\right]_{i,i}+2NQ \ln (\sigma).
\end{aligned}
\end{equation}

The new loss function enables probabilistic forecasting by first drawing samples $\Tilde{\boldsymbol{E}_t}$ from $\mathcal{N}(\boldsymbol{0}_{NQ},\boldsymbol{\Sigma})$ and subsequently transforming these samples into the target variable through the following process:
\begin{equation}\label{eqn:sampling}
\Tilde{\boldsymbol{Y}}_t = \hat{\boldsymbol{Y}}_t + \Tilde{\boldsymbol{E}}_t,
\end{equation}
where $\hat{\boldsymbol{Y}}_t$ is the predicted mean as defined in Eq.~\eqref{eqn:mean_pred}. Since the covariance matrix \(\boldsymbol{\Sigma}\) captures the contemporaneous correlations in $\boldsymbol{E}_t$, we expect this approach to improve uncertainty quantification in probabilistic forecasting.

\section{Experiment}\label{sec:exp}

To evaluate the effectiveness of our method, we conduct experiments on two traffic datasets: PEMSD7 (M) and PEMS08. PEMSD7 (M) is a highway traffic speed dataset initially employed in STGCN \citep{yu2017spatio}. PEMS08 is a highway traffic flow dataset originally utilized in ASTGCN \citep{guo2019attention}. We follow the identical data processing procedures outlined in the original studies. For PEMS08, we allocate 60\% of the data for training, 20\% for validation, and the remaining 20\% for testing. For PEMSD7 (M), we use a 7:1:2 split. Across all datasets, we normalize data using z-score transformation based on training set statistics. Missing values are excluded from both training and testing. Table~\ref{tab:datasets} summarizes dataset details.

\begin{table}[t]
\caption{Dataset description.}\label{tab:datasets}
    \centering
    \begin{tabular}{lccccc}
        \toprule
        Datasets & Type & \#Nodes & \#Time Steps & Resolution\\
        \midrule
        PEMSD7 (M) & Speed & 228 & 12,672 & 5 min\\
        PEMS08 & Flow & 170 & 17,856 & 5 min\\
        \bottomrule
    \end{tabular}
\end{table}

\subsection{Base Models}

In addition to the models originally designed for the selected datasets, we include three additional models as base models $f_{\theta}(\cdot)$ to diversify the range of architectures evaluated in our approach. Most existing STF frameworks combine GNNs with sequential models to effectively model traffic data. The choice of sequential models encompasses RNN-based methods (e.g., AGCRN \citep{bai2020adaptive}), TCN-based methods (e.g., STGCN \citep{yu2017spatio} and Graph WaveNet \citep{wu2019graph}), and attention-based methods (e.g., ASTGCN \citep{guo2019attention}). Furthermore, we incorporate a SOTA method, D\textsuperscript{2}STGNN \citep{shao2022decoupled}, which is highlighted in the LargeST benchmark \citep{liu2024largest}. The implementation of the proposed methods is publicly available at: \href{https://github.com/rottenivy/dynamic-regression}{\texttt{https://github.com/rottenivy/dynamic-regression}}.

\subsection{Experimental Setup}

Our experimental setup involves a computer environment with an Intel(R) Xeon(R) CPU E5-2698 v4 @ 2.20 GHz and four NVIDIA Tesla V100 GPUs. Base models were implemented using either the original source code or their PyTorch versions. All models were configured to employ 12 historical observation steps ($P=12$) for predicting 12 future steps ($Q=12$). During training, the Adam optimizer was utilized with an initial learning rate of 0.001 and a weight decay of 0.0001. To prevent over-fitting, early stopping was employed when the validation loss showed consistent increase over 15 epochs. All metrics are averaged over three independent runs.

For the error AR process, we need to determine the lag length $\Delta $. Since traffic data exhibits strong local correlation and seasonality, we mainly consider the correlation between the current residual and its 1) most recent available predecessor ($\Delta = Q$); 2) counterpart one day apart ($\Delta = \text{1 day}$); and 3) counterpart one week apart ($\Delta  = \text{1 week}$). Although the use of seasonal lags is a common practice in time series modeling across various domains, such as in Seasonal ARIMA models \citep{hyndman2018forecasting}, the explicit incorporation of seasonal error correlations is novel in the context of prior related works addressing correlated errors in traffic data \citep{sun2021adjusting,kim2022residual}. We determine the optimal $\Delta $ based on validation loss for each model and dataset.

The final loss function of our method is composed of two parts:
\begin{equation}
\label{eq:loss_func}
\mathcal{L} = \mathcal{L}_{\text{nll}} + \mathcal{L}_{\text{reg}},
\end{equation}
where $\mathcal{L}_{\text{reg}}$ is the $\ell_1$ regularization in Eq.~\eqref{eqn:loss} and $\mathcal{L}_{\text{nll}}$ is the non-isotropic loss function in Eq.~\eqref{eq:nll_mnd}. For comparative analyses, we also train the base model without our DR framework, utilizing an isotropic loss function and excluding the error AR process, i.e., $\mathcal{L}\propto \boldsymbol{\eta}_t^\top\boldsymbol{\eta}_t$ by assuming $\boldsymbol{\eta}_t=\operatorname{vec}(\boldsymbol{R}_t) \sim \mathcal{N}(\mathbf{0}, \sigma^2\mathbf{I}_{NQ})$. This is equivalent to using MSE as the loss function. In comparison to the base model trained without our method, we introduce four additional trainable parameters, i.e., regression coefficients $\boldsymbol{A}\in\mathbb{R}^{N\times N}$ and $\boldsymbol{B}\in\mathbb{R}^{Q\times Q}$ for the error AR process, and two low-rank matrices $\boldsymbol{L}_N \in \mathbb{R}^{N \times R_n}$ and $\boldsymbol{L}_Q \in \mathbb{R}^{Q \times R_q}$ for the error covariance learning. We use \(R_n = N\) and \(R_q = Q\) for the initial comparison; however, sensitivity analysis will be conducted to assess the impact of these rank parameters. However, the additional parameters are negligible compared to the overall size of the base model.

We use several evaluation metrics for validating the effectiveness of our method, namely, root relative mean squared error (RRMSE) \citep{bai2020adaptive,lai2018modeling,song2021dual,shih2019temporal}, empirical Continuous Ranked Probability Score (CRPS) \citep{gneiting2007strictly}, and quantile loss ($\rho$-risk) \citep{salinas2020deepar}.

$\operatorname{RRMSE}$ is mainly used to assess the accuracy of point forecasts of the models, as it directly evaluates the predicted mean $\hat{\boldsymbol{Y}}_t$:
\begin{equation}
    \operatorname{RRMSE} = \frac{\sqrt{\sum_t \|\boldsymbol{Y}_t - \hat{\boldsymbol{Y}}_t\|_F^2}}{\sqrt{\sum_t \|\boldsymbol{Y}_t - \bar{\boldsymbol{Y}}_t\|_F^2}},
\end{equation}
where $\bar{\boldsymbol{Y}}_t$ is the mean value of the testing dataset. Lower $\operatorname{RRMSE}$ values indicate better accuracy. As it solely evaluates the predicted mean, $\operatorname{RRMSE}$ primarily reflects the effectiveness of the error AR module.

\(\operatorname{CRPS}\) is a widely used quantitative measure to evaluate the quality of probabilistic forecasts when the observation is a scalar \citep{salinas2019high,rasul2020multivariate,rasul2021autoregressive,drouin2022tactis,ashok2023tactis}. In this work, we use the kernel representation form of the \(\operatorname{CRPS}\) \citep{gneiting2007strictly}:
\begin{equation}\label{eqn:crps}
    \operatorname{CRPS}\left(F, z\right)=\operatorname{E}_F\lvert Z-z \rvert-\frac{1}{2} \operatorname{E}_F\lvert Z-Z^{\prime} \rvert,
\end{equation}
where $z$ is the observation, $F$ is the CDF of the predicted variable, $Z$ and $Z^{\prime}$ are independent copies of a set of prediction samples associated with the distribution $F$. Lower $\operatorname{CRPS}$ values indicate better agreement between predicted and observed distributions. It can be shown that \(\operatorname{CRPS}\) inherently accounts for the calibration and sharpness of probabilistic forecasts:
\begin{itemize}
    \item Overconfident (Too Sharp): Forecasts that are too narrow will have larger squared differences where they fail to encompass the observed value, leading to a higher \(\operatorname{CRPS}\) (the first term in Eq.~\eqref{eqn:crps}).
    \item Underconfident (Too Wide): Forecasts that are too wide will have persistent deviations over a broad range of values, also leading to a higher \(\operatorname{CRPS}\) (the second term in Eq.~\eqref{eqn:crps}).
\end{itemize}

We compute \(\operatorname{CRPS}\) at each spatial location and each forecast step by drawing samples from Eq.~\eqref{eqn:sampling}. The scores are summarized by first summing across the entire testing horizon for all time series and then normalizing by the sum of the corresponding ground truth values.

The quantile loss $L_\rho\left(z, \hat{Z}^\rho\right)$ is used to quantify the accuracy of a given quantile, denoted by $\rho$, from the predictive distribution:
\begin{equation}
\label{eq:q_loss}
    L_\rho\left(z, \hat{Z}^\rho\right)=2(\hat{Z}^\rho-z)\left((1-\rho) \mathrm{I}_{\hat{Z}^\rho>z}-\rho \mathrm{I}_{\hat{Z}^\rho \leq z}\right),
\end{equation}
where $\mathrm{I}$ is a binary indicator function that equals 1 when the condition is met, $\hat{Z}^\rho$ represents the predicted $\rho$-quantile, and $z$ represents the ground truth value. We summarize the quantile losses for the testing set across all time series segments by evaluating a normalized summation of these losses: $\left(\sum_t L_\rho\left(z_t, \hat{Z}_t^\rho\right)\right) /\left(\sum_t z_t\right)$. In this paper, we evaluate the $0.5$-risk, $0.75$-risk, and the $0.9$-risk, calculated from the corresponding quantiles of the empirical predictive distribution. We drew 100 prediction samples to form the empirical predictive distribution used to calculate the probabilistic scores, specifically the \(\operatorname{CRPS}\) and the $\rho$-risks.

\subsection{Experimental Results}

\begin{figure}[htbp]
  \centering
  \includegraphics[width=0.75\textwidth, interpolate=false]{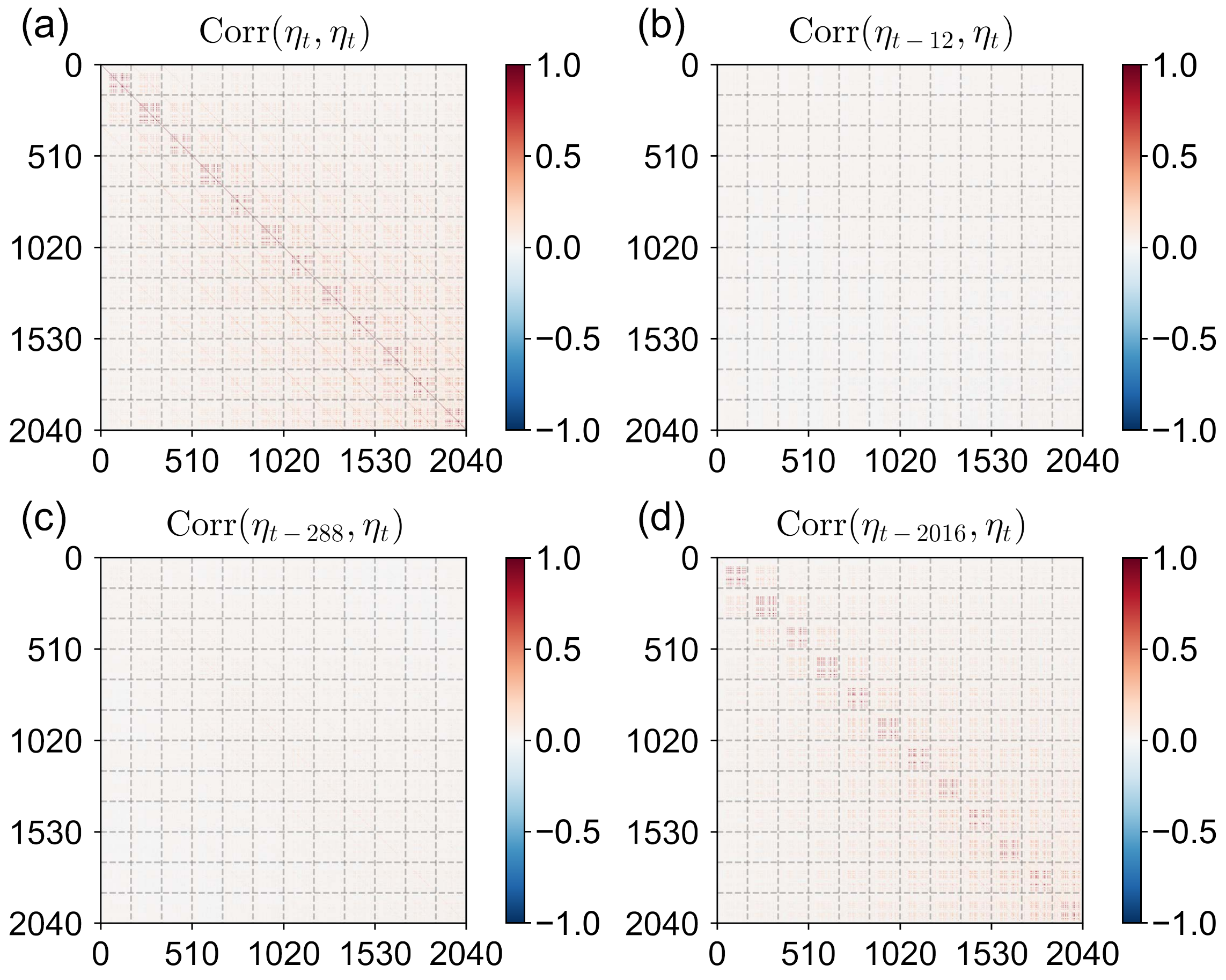}
  \caption{Correlation matrices of residuals obtained from the original Graph WaveNet model on the PEMS08 dataset. From (a) to (d): Contemporaneous correlations, Cross-correlations at lag 12 (i.e., most recent), lag 288 (i.e., 1 day), lag 2016 (i.e., 1 week).}
  \label{fig:fig2_gwave_pm08_res}
\end{figure}

We first analyze cross-correlation and contemporaneous correlation in residuals from a base model without our method. We calculate two types of correlation: $\operatorname{Corr}(\boldsymbol{\eta}_t,\boldsymbol{\eta}_t)$ and $\operatorname{Corr}(\boldsymbol{\eta}_{t-\Delta }, \boldsymbol{\eta}_t)$. $\operatorname{Corr}(\boldsymbol{\eta}_t,\boldsymbol{\eta}_t)$ is the contemporaneous correlation of the variables in $\boldsymbol{\eta}_t$, while $\operatorname{Corr}(\boldsymbol{\eta}_{t-\Delta }, \boldsymbol{\eta}_t)$ is the cross-correlation at lag $\Delta$. If the error independently follows an identical isotropic distribution, $\operatorname{Corr}(\boldsymbol{\eta}_t,\boldsymbol{\eta}_t)$ should be an identity matrix and $\operatorname{Corr}(\boldsymbol{\eta}_{t-\Delta }, \boldsymbol{\eta}_t)$ should be a zero matrix. In Figure~\ref{fig:fig2_gwave_pm08_res}, we present the residual correlation matrices of PEMS08 using the results from Graph WaveNet. We observe that $\operatorname{Corr}(\boldsymbol{\eta}_t,\boldsymbol{\eta}_t)$ is not diagonal, suggesting there exists spatial and across-step correlations in the residuals. In terms of $\operatorname{Corr}(\boldsymbol{\eta}_{t-\Delta},\boldsymbol{\eta}_t)$, we examine different values including $\Delta=12$ (1 hour), $\Delta=288$ (1 day), and $\Delta=2016$ (1 week). Interestingly, we find that $\Delta=2016$ exhibits the strongest cross-correlation patterns, while correlations with $\Delta=12$ and $\Delta=288$ are weak. We believe this is mainly due to the fact that traffic flow is heavily determined by travel demand, which often exhibits prominent weekly periodicity. We selected the optimal value of $\Delta$ for each dataset based on the validation loss. Specifically, in our study, we employed $\Delta=12$ for PEMSD7 (M) and $\Delta=2016$ for PEMS08.

\subsubsection{Overall Comparison}

\begin{table*}[htbp]
\caption{Impact of the proposed method on various models and datasets.}\label{tab:benchmarks}
\centering
\begin{tabular}{ccccccc}
\toprule
Data & Model & $\operatorname{RRMSE}$ & $\operatorname{CRPS}$ & $0.5$-risk & $0.75$-risk & $0.9$-risk\\
\midrule
\multirow{9}{*}{\rotatebox[origin=c]{90}{PEMSD7 (M)}}
 & STGCN            & 0.4801          & 0.0566  & 0.0699 & 0.0591          & 0.0385          \\
 &   \quad\quad + DR     & \textbf{0.4501} & \textbf{0.0544}          & \textbf{0.0691}          & \textbf{0.0575} & \textbf{0.0363} \\
 \cmidrule(lr){2-7}
 & Graph WaveNet    & 0.3676          & 0.0427          & 0.05            & 0.0461          & 0.0308          \\
 &   \quad\quad + DR     & \textbf{0.3622} & \textbf{0.0389} & \textbf{0.0478} & \textbf{0.0409} & \textbf{0.0268} \\
 \cmidrule(lr){2-7}
 & AGCRN            & \textbf{0.375} & 0.0435          & 0.0509          & 0.0471          & 0.0316          \\
 &   \quad\quad + DR     & 0.3838         & \textbf{0.0406} & \textbf{0.0499} & \textbf{0.0427} & \textbf{0.0289} \\
 \cmidrule(lr){2-7}
  & D\textsuperscript{2}STGNN            & 0.3639      & 0.0416          & 0.0489          & 0.0452          & 0.0305          \\
   & \quad\quad + DR      & \textbf{0.3592}         & \textbf{0.0391} & \textbf{0.0481} & \textbf{0.0407} & \textbf{0.0271} \\
 \cmidrule(lr){2-7}
 & Avg. rel. impr.  & 1.67\%         & 6.37\%          & 2.29\%          & 8.32\%          & 9.60\% \\
\midrule
\multirow{9}{*}{\rotatebox[origin=c]{90}{PEMS08}}
 & ASTGCN           & 0.1955          & 0.0649          & 0.0841          & 0.0756          & 0.0476          \\
 &   \quad\quad + DR     & \textbf{0.1838} & \textbf{0.0566} & \textbf{0.0758} & \textbf{0.0638} & \textbf{0.0397} \\
 \cmidrule(lr){2-7}
 & Graph WaveNet    & 0.1707          & 0.0544          & 0.0698          & 0.0614          & 0.0389          \\
 &   \quad\quad + DR     & \textbf{0.1639} & \textbf{0.0465} & \textbf{0.0619} & \textbf{0.0517} & \textbf{0.0321} \\
 \cmidrule(lr){2-7}
 & AGCRN            & 0.1789          & 0.0579          & 0.0746          & 0.0665          & 0.0422          \\
 &   \quad\quad + DR     & \textbf{0.1739} & \textbf{0.0526} & \textbf{0.0698} & \textbf{0.0602} & \textbf{0.0392} \\
\cmidrule(lr){2-7}
 & D\textsuperscript{2}STGNN            & 0.1678          & 0.0537          & 0.0683          & 0.0615          & 0.0388          \\
  & \quad\quad + DR      & \textbf{0.1665}         & \textbf{0.0481} & \textbf{0.0645} & \textbf{0.0545} & \textbf{0.0331} \\
 \cmidrule(lr){2-7}
 & Avg. rel. impr.   & 3.38\%         & 11.72\%         & 8.30\%          & 13.07\%         & 13.97\% \\
\bottomrule
\end{tabular}
\end{table*}

\begin{figure}[htbp]
  \centering
  \includegraphics[width=0.75\textwidth, interpolate=false]{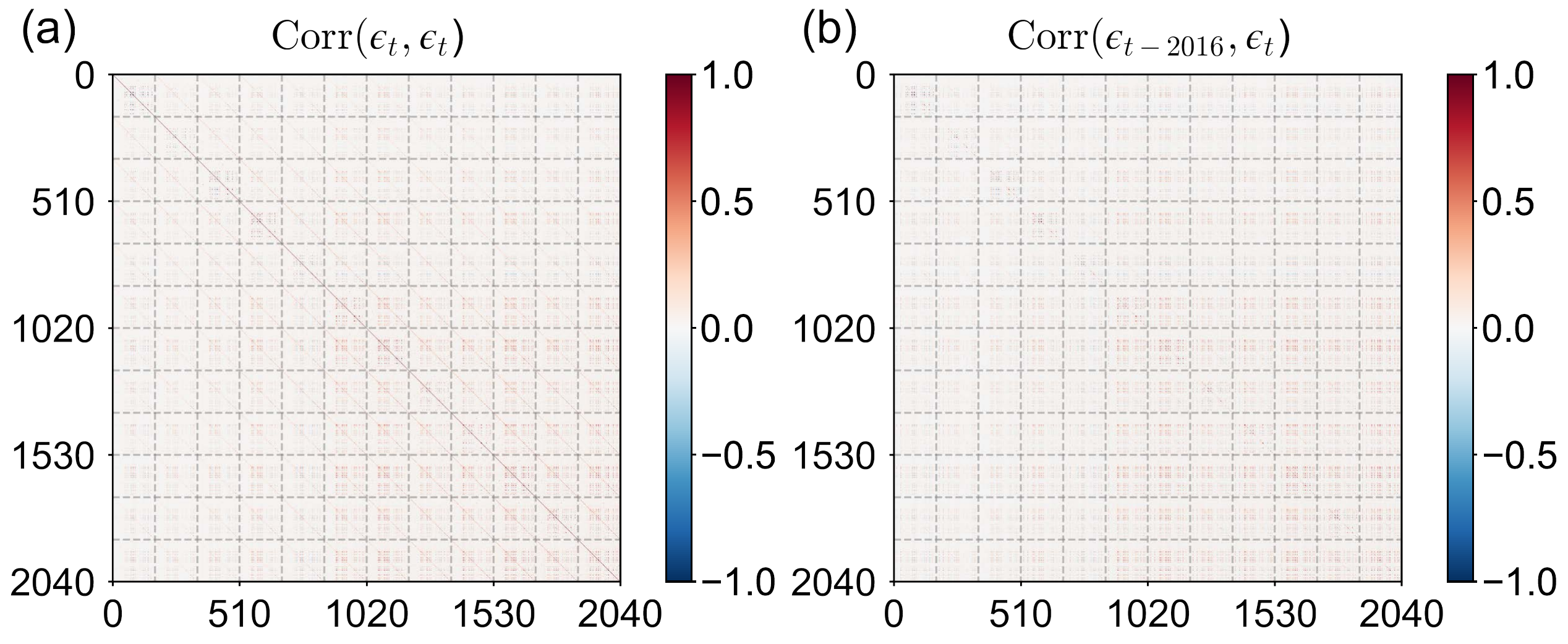}
  \caption{Correlation matrices of residuals obtained from the Graph WaveNet model on the PEMS08 dataset after applying our method. (a) Contemporaneous correlations; (b) Cross-correlations at lag 2016 (i.e., one week).}
  \label{fig:fig3_gwave_dr_pm08_res}
\end{figure}

Table \ref{tab:benchmarks} offers a comprehensive overview of our approach's performance, encompassing various deep learning models and datasets assessed through diverse metrics. The chosen metrics shed light on multiple facets of model effectiveness. Notably, our method demonstrates improvements of $1.67\%$ and $3.38\%$ concerning $\operatorname{RRMSE}$ for PEMSD7 (M) and PEMS08, respectively. Since $\operatorname{RRMSE}$ evaluates only the mean prediction $\hat{\boldsymbol{Y}}_t$, it directly measures the effectiveness of the error AR module. The comparatively modest improvement in $\operatorname{RRMSE}$ for PEMSD7 (M) compared to PEMS08 implies that the base models already exhibit commendable performance on this specific dataset.

The $\operatorname{CRPS}$ and $\rho$-risk are probabilistic scores employed to assess the efficacy of probabilistic forecasting. Notably, our method consistently demonstrates superior performance across nearly all scenarios, as reflected in these probabilistic scores. Specifically, $\operatorname{CRPS}$ evaluates the overall accuracy of the predictive distribution. Our approach yields average improvements of $6.37\%$ and $11.72\%$ on PEMSD7 (M) and PEMS08, respectively. The $\rho$-risk is assessed at different quantiles ($\rho=0.5, 0.75, 0.9$) in this study. In accordance with Eq.~\eqref{eq:q_loss}, the $0.5$-risk essentially equates to the MAE between the actual value and the 0.5-quantile of the predictive distribution. Conversely, the $0.75$-risk and $0.9$-risk apply less penalty when the associated quantiles surpass the actual value. For instance, the 0.9-quantile will be heavily penalized if it falls below the actual value. Our method significantly improves the $0.9$-risk, where improvements amount to $9.60\%$ and $13.97\%$ on PEMSD7 (M) and PEMS08, respectively.

Figure~\ref{fig:fig3_gwave_dr_pm08_res} visually demonstrates a significant reduction in both contemporaneous correlation and cross-correlation, as compared to Figure~\ref{fig:fig2_gwave_pm08_res}. It is essential to note that the reduction in both contemporaneous correlation and cross-correlation can be attributed to the error AR module, as more accurate predictions guide the error $\boldsymbol{\epsilon}_t$ towards isotropy. Simultaneously, spatiotemporal covariance learning has the effect of flexibly regularizing the error by allowing the existence of contemporaneous correlation.

\subsubsection{Ablation Studies}

In this section, we conduct ablation studies and sensitivity analyses to assess the contribution and effectiveness of each component within the DR framework.

\noindent\textbf{Component Ablations in the DR Framework.} Table~\ref{tab:abla_dr} shows that the error AR process primarily enhances the mean prediction, as indicated by the significant improvement in $\operatorname{RRMSE}$ without $\mathcal{L}_{\text{nll}}$. On the other hand, the non-isotropic loss alone (w/o AR) provides a mild enhancement to the model in terms of $\operatorname{RRMSE}$ since it only regularizes the loss with a more flexible assumption. However, the full covariance learning appears to be beneficial when performing probabilistic forecasting, as indicated by the improvement seen in the $0.75$-risk and $0.9$-risk.

\noindent\textbf{Sensitivity Analysis of Rank Parameters.} Since \(Q=12\) is already a small value, we fix \(R_q=Q=12\) in this study and perform ablations on \(R_n\). Table~\ref{tab:abla_rank} shows that reducing \(R_n\) degrades probabilistic performance, as seen in $\operatorname{CRPS}$ and $\rho$-risk metrics. This decline is attributed to the reduced ability to effectively model the covariance. Due to the probabilistic nature of this component, its impact on $\operatorname{RRMSE}$ appears minimal and random.

\noindent\textbf{Performance Limits of the DR Framework.} To understand the performance limits of the proposed framework, we also investigate the predictive power of using only a bi-linear AR model similar to Eq.~\eqref{eqn:matrix_ar1}, with \(\boldsymbol{R}_{t}\) and \(\boldsymbol{R}_{t-\Delta}\) replaced by \(\boldsymbol{Y}_{t}\) and \(\boldsymbol{Y}_{t-\Delta}\). Table~\ref{tab:abla_dl_dr} shows that the DL model alone outperforms the AR model regardless of the loss function used. This is because the AR model is too simple to effectively capture the traffic patterns. Furthermore, the performance of the AR model improves significantly when using a higher order AR process, highlighting its dependency on richer temporal dynamics. Lastly, the non-isotropic loss consistently enhances performance when applied to the same model, demonstrating its effectiveness in refining predictions.

Neither the DL model nor the AR model alone is able to outperform our proposed method. Although the DL model can theoretically capture all spatiotemporal correlations, making residuals i.i.d., achieving this in practice is difficult. It requires careful curation of spatiotemporal features and substantial effort in model design and optimization. In contrast, our method provides a model-agnostic alternative that enhances prediction performance by leveraging the statistical properties of the error term, offering a more practical and efficient solution.

\begin{table}[htbp]
\caption{Ablation studies on DR components using Graph WaveNet and the PEMS08 dataset.}\label{tab:abla_dr}
\centering
\begin{tabular}{cccccccc}
\toprule
 Model & $\operatorname{RRMSE}$ & $\operatorname{CRPS}$ & $0.5$-risk & $0.75$-risk & $0.9$-risk\\
\midrule
 Graph WaveNet            & 0.1707             & 0.0544              & 0.0698            & 0.0614           & 0.0389          \\
 \multicolumn{1}{l}{\quad\quad\quad\quad + DR (w/o AR)}        & 0.1702             & 0.0543              & 0.0698            & 0.0605           & 0.0384 \\
 \multicolumn{1}{l}{\quad\quad\quad\quad + DR (w/o $\mathcal{L}_{\text{nll}}$)}     & \textbf{0.1628}              & 0.0524            & 0.0650            & 0.0606           & 0.0384 \\
 \multicolumn{1}{l}{\quad\quad\quad\quad + DR}               & 0.1639    & \textbf{0.0465}     & \textbf{0.0619}   & \textbf{0.0517}  & \textbf{0.0321} \\
\bottomrule
\end{tabular}
\end{table}

\begin{table}[htbp]
\caption{Sensitivity analysis on the rank parameters \(R_n\) using Graph WaveNet and the PEMS08 dataset.}\label{tab:abla_rank}
\centering
\begin{tabular}{ccccccccc}
\toprule
 Model & $\operatorname{RRMSE}$ & $\operatorname{CRPS}$ & $0.5$-risk & $0.75$-risk & $0.9$-risk\\
\midrule
 Graph WaveNet                          & 0.1707 & 0.0544 & 0.0698 & 0.0614 & 0.0389 \\
 \multicolumn{1}{l}{\quad\quad\quad\quad + DR (\(R_n=N\))}       & 0.1639 & 0.0465 & 0.0619 & \textbf{0.0517} & \textbf{0.0321} \\
 \multicolumn{1}{l}{\quad\quad\quad\quad + DR (\(R_n=120\))}     & \textbf{0.1612} & \textbf{0.0464} & \textbf{0.0611} & 0.0525 & 0.0322 \\
 \multicolumn{1}{l}{\quad\quad\quad\quad + DR (\(R_n=80\))}      & 0.1626 & 0.0482 & 0.0628 & 0.0551 & 0.0338 \\
 \multicolumn{1}{l}{\quad\quad\quad\quad + DR (\(R_n=40\))}      & 0.1619 & 0.0492 & 0.0633 & 0.0565 & 0.0349 \\
 \multicolumn{1}{l}{\quad\quad\quad\quad + DR (\(R_n=10\))}      & 0.1623 & 0.0506 & 0.064  & 0.0581 & 0.0366 \\
\bottomrule
\end{tabular}
\end{table}

\begin{table}[htbp]
\caption{Empirical performance limits of our method using Graph WaveNet and the PEMS08 dataset.}\label{tab:abla_dl_dr}
\centering
\begin{tabular}{ccccccccc}
\toprule
 Loss & Model & $\operatorname{RRMSE}$ & $\operatorname{CRPS}$ & $0.5$-risk & $0.75$-risk & $0.9$-risk\\
\midrule
 \multirow{3}{*}{isotropic} & DL only  & 0.1707 & 0.0544 & 0.0698 & 0.0614 & 0.0389 \\
 & AR \((p=1)\) only        & 0.2838 & 0.0908 & 0.1102 & 0.1081 & 0.0705 \\
 & AR \((p=2)\) only        & 0.2666 & 0.0863 & 0.106  & 0.1007 & 0.0646 \\
 & AR \((p=3)\) only        & 0.2054 & 0.0677 & 0.0853 & 0.0794 & 0.0505 \\
\cmidrule(lr){1-7}
 \multirow{3}{*}{non-isotropic} & DL only & 0.1702 & 0.0543 & 0.0698 & 0.0605 & 0.0384 \\
 & AR \((p=1)\) only        & 0.2599 & 0.0753 & 0.0984 & 0.0834 & 0.0546 \\
 & AR \((p=2)\) only        & 0.2439 & 0.0709 & 0.093  & 0.0802 & 0.0532 \\
 & AR \((p=3)\) only        & 0.1948 & 0.0596 & 0.0778 & 0.0688 & 0.0445 \\
\cmidrule(lr){1-7}
 non-isotropic & DR               & \textbf{0.1639}  & \textbf{0.0465}  & \textbf{0.0619}  & \textbf{0.0517}  & \textbf{0.0321} \\
\bottomrule
\end{tabular}
\end{table}

\subsubsection{Model Interpretation}

\begin{figure}[htbp]
  \centering
  \includegraphics[width=0.65\textwidth, interpolate=false]{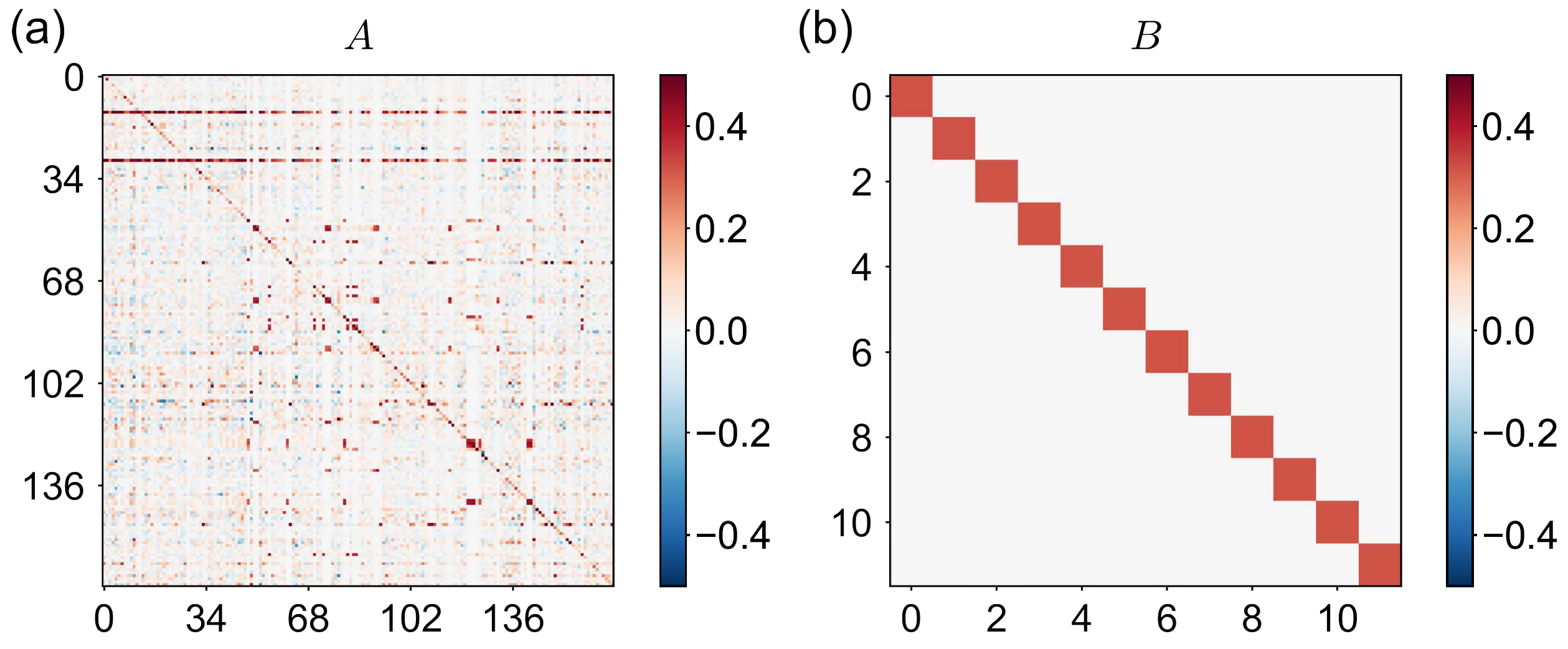}
  \caption{Learned AR coefficient matrices $\boldsymbol{A}$ and $\boldsymbol{B}$ for the Graph WaveNet model on the PEMS08 dataset.}
  \label{fig:fig4_gwave_pm08_ab}
\end{figure}

\begin{figure}[htbp]
  \centering
  \includegraphics[width=0.65\textwidth, interpolate=false]{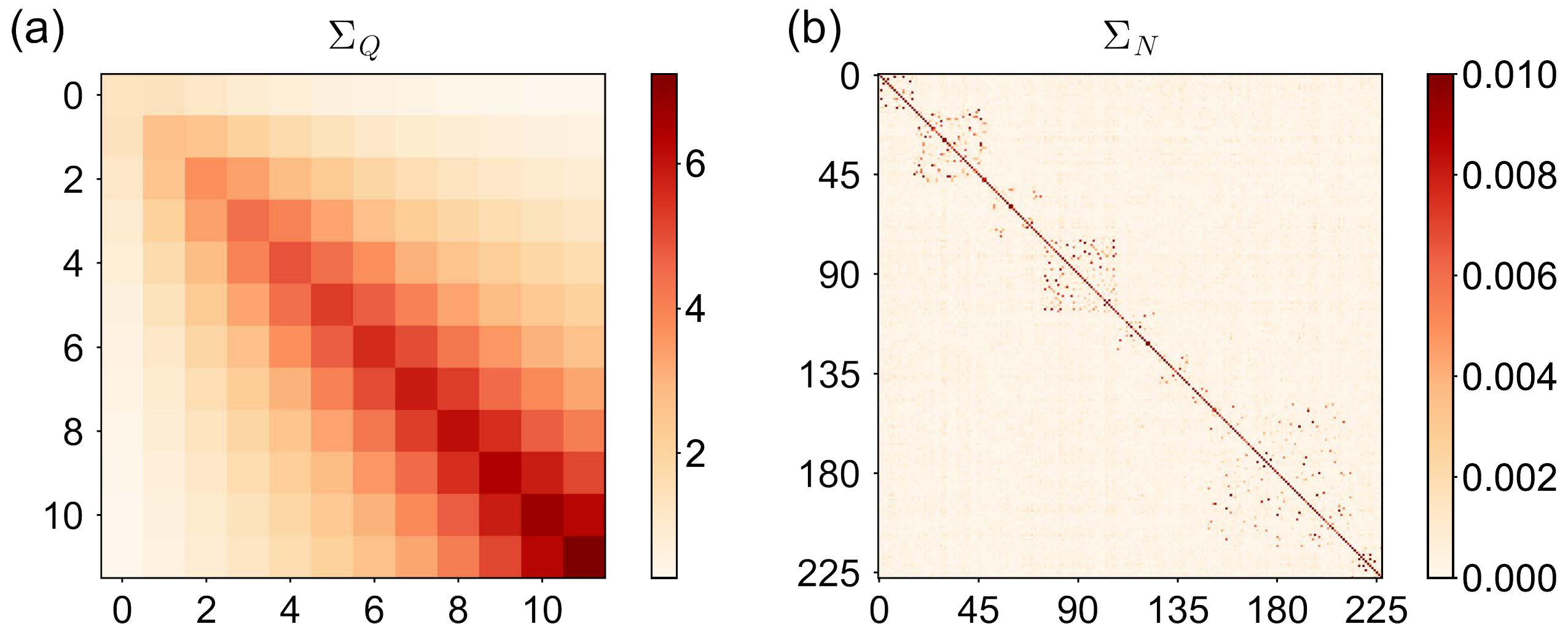}
  \caption{Learned covariance matrices $\boldsymbol{\Sigma}_Q$ and $\boldsymbol{\Sigma}_N$ for the Graph WaveNet model on the PEMSD7(M) dataset.}
  \label{fig:fig5_gwave_pm7m_cov}
\end{figure}

\begin{figure}[htbp]
  \centering
  \includegraphics[width=0.9\textwidth, interpolate=false]{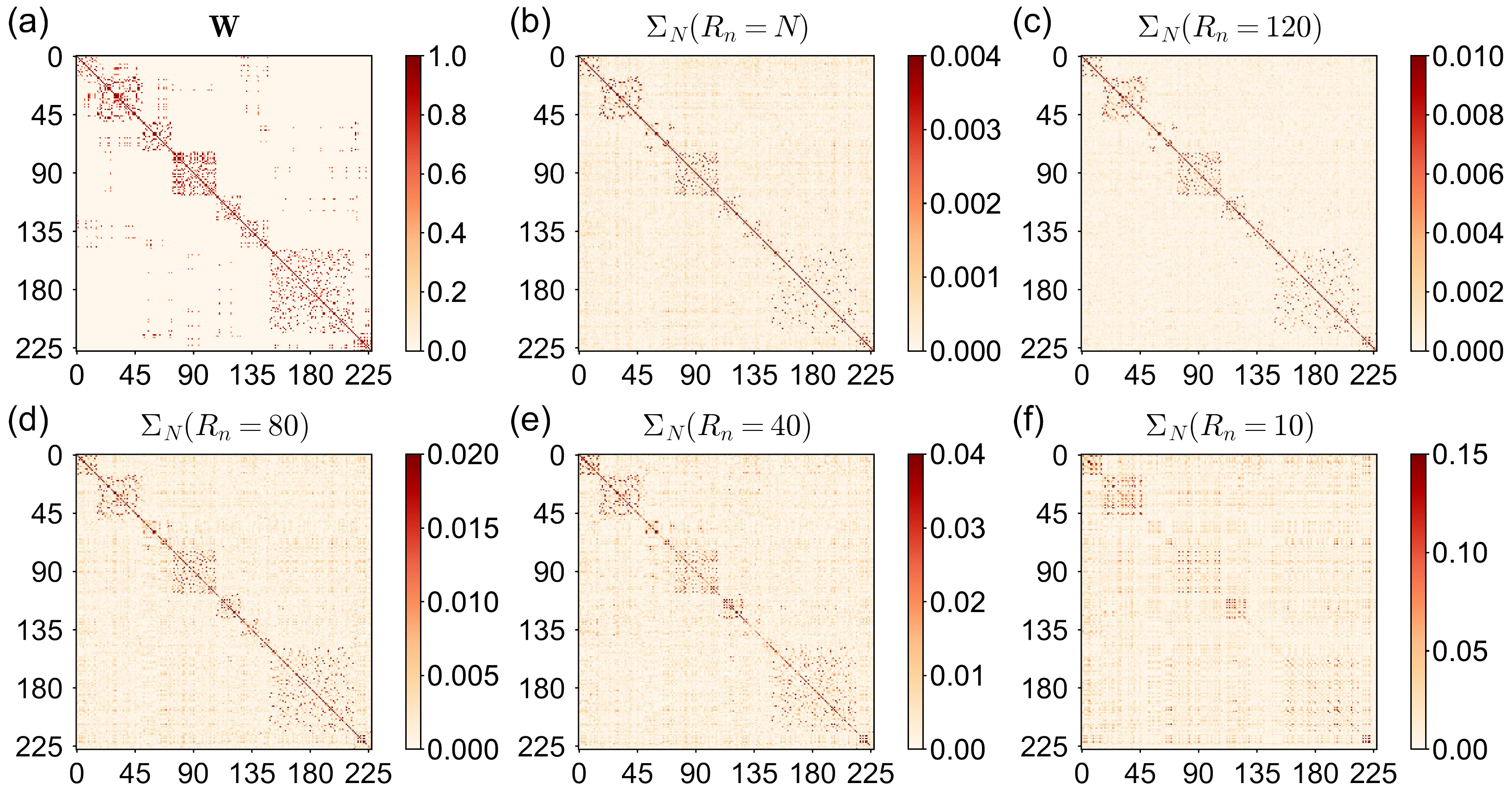}
  \caption{Comparison of the adjacency matrix and learned covariance matrices \(\boldsymbol{\Sigma}_N\) with different rank parameters \(R_n\) using Graph WaveNet and the PEMSD7(M) dataset. Values in \(\boldsymbol{\Sigma}_N\) are clipped to a specific range to enhance the visibility of the patterns.}
  \label{fig:compare_rank_cov}
\end{figure}

We analyze coefficient matrices ($\boldsymbol{A}$ and $\boldsymbol{B}$) and covariance matrices ($\boldsymbol{\Sigma}_N$ and $\boldsymbol{\Sigma}_Q$) to interpret the model. Figure~\ref{fig:fig4_gwave_pm08_ab} showcases the coefficient matrices of the error AR module in Graph WaveNet on PEMS08. In Figure~\ref{fig:fig4_gwave_pm08_ab} (a), the correlations between residuals from different spatial locations are depicted. The prominent diagonal reveals that most spatial locations exhibit strong self-correlations, indicating that the residuals of the base DL model are autocorrelated. Intriguingly, certain locations display strong correlations with residuals from other locations, indicating that the spatial dependency has not been fully explored by the DL model. Figure~\ref{fig:fig4_gwave_pm08_ab} (b) further highlights a pronounced diagonal in matrix $\boldsymbol{B}$. Given that $\boldsymbol{B}$ signifies the column effect of the past residual $\boldsymbol{R}_{t-\Delta }$ on $\boldsymbol{R}_{t}$, the current residual showcases the highest correlation with the past residual at the corresponding forecasting step.

Figure~\ref{fig:fig5_gwave_pm7m_cov} presents the learned error covariance matrices obtained for Graph WaveNet on PEMSD7 (M). The temporal covariance matrix \(\boldsymbol{\Sigma}_Q\) (Figure~\ref{fig:fig5_gwave_pm7m_cov} (a)) encapsulates the covariance across the prediction horizon. We observe that the diagonal elements of \(\boldsymbol{\Sigma}_Q\) progressively expand with each prediction step, an intuitively rational behavior for multistep prediction tasks, where the uncertainty of prediction increases as we predict further into the future. This knowledge can be translated into probabilistic forecasting, where the prediction interval increases along the prediction horizon, providing additional information for the downstream decision-making process. Furthermore, the visualization reveals a discernible propagation of errors in the extended prediction period, evidenced by the elevated covariance between consecutive prediction steps. This suggests DL models could be improved to mitigate error propagation in multistep predictions. For example, one could leverage the observed residuals to calibrate next-step predictions in an autoregressive manner \citep{zheng2024better,zheng2024multivariate}.

Figure~\ref{fig:fig5_gwave_pm7m_cov}~(b) illustrates the spatial covariance matrix \(\boldsymbol{\Sigma}_N\), which captures the statistical dependencies among the residuals across different spatial locations. Although \(\boldsymbol{\Sigma}_N\) may exhibit some similarities to the adjacency matrix commonly used in GNNs  (Figure~\ref{fig:compare_rank_cov}~(a)), its role and interpretation are fundamentally distinct. Whereas the adjacency matrix encodes predefined or learned spatial dependencies, the covariance matrix captures error correlations that GNNs miss. These correlations may stem from factors not explicitly represented in the adjacency matrix, such as correlated noise or external influences.

The low-rank parameterization of \(\boldsymbol{\Sigma}_Q\) and \(\boldsymbol{\Sigma}_N\) also offers computational benefits. For example, as \(R_n\) decreases, the model demonstrates robustness by retaining its ability to recover covariance patterns represented by the full-rank parameterization (i.e., \(R_n=N\)), as shown in Figure~\ref{fig:compare_rank_cov}~(b-e). This indicates the model can capture essential spatial dependencies even with reduced parameterization. However, when \(R_n\) is reduced to a very small value of 10 (Figure~\ref{fig:compare_rank_cov}~(f)), the model begins to lose its capacity to recover intricate covariance patterns. Moreover, this parameterization represents each spatial location and forecasting step through \(R_n\) and \(R_q\) latent factors, respectively. The covariance structures induced by these factors are naturally expressed as \(\boldsymbol{L}_N\boldsymbol{L}_N^\top\) for spatial correlations and \(\boldsymbol{L}_Q\boldsymbol{L}_Q^\top\) for temporal correlations.

The fully learnable design of the proposed method serves as a robust enhancement to existing DL models. In cases of model misspecification, our method can theoretically reduce to the original implementation. As shown in Eq.~\eqref{eqn:dr_model} and Eq.~\eqref{eq:nll_mnd}, the model becomes identical to the original implementation with an isotropic loss when \(\boldsymbol{A}\)/\(\boldsymbol{B}\) and \(\boldsymbol{\Sigma}_Q\)/\(\boldsymbol{\Sigma}_N\) are zero matrices. Thus, our method generalizes the conventional approach, allowing the model to dynamically adapt to the error generation and propagation process during training. Furthermore, Figure~\ref{fig:abla_ar_order} indicates that the non-isotropic loss is more robust than the isotropic loss when there is misspecification in the error AR process.

\subsubsection{Validation of Assumptions}

In this section, we present experimental results that validate the two assumptions (Assumption \ref{asmp1} and Assumption \ref{asmp2}) stated in Section \ref{sec:method}.

\begin{figure}[htbp]
  \centering
  \includegraphics[width=0.8\textwidth, interpolate=false]{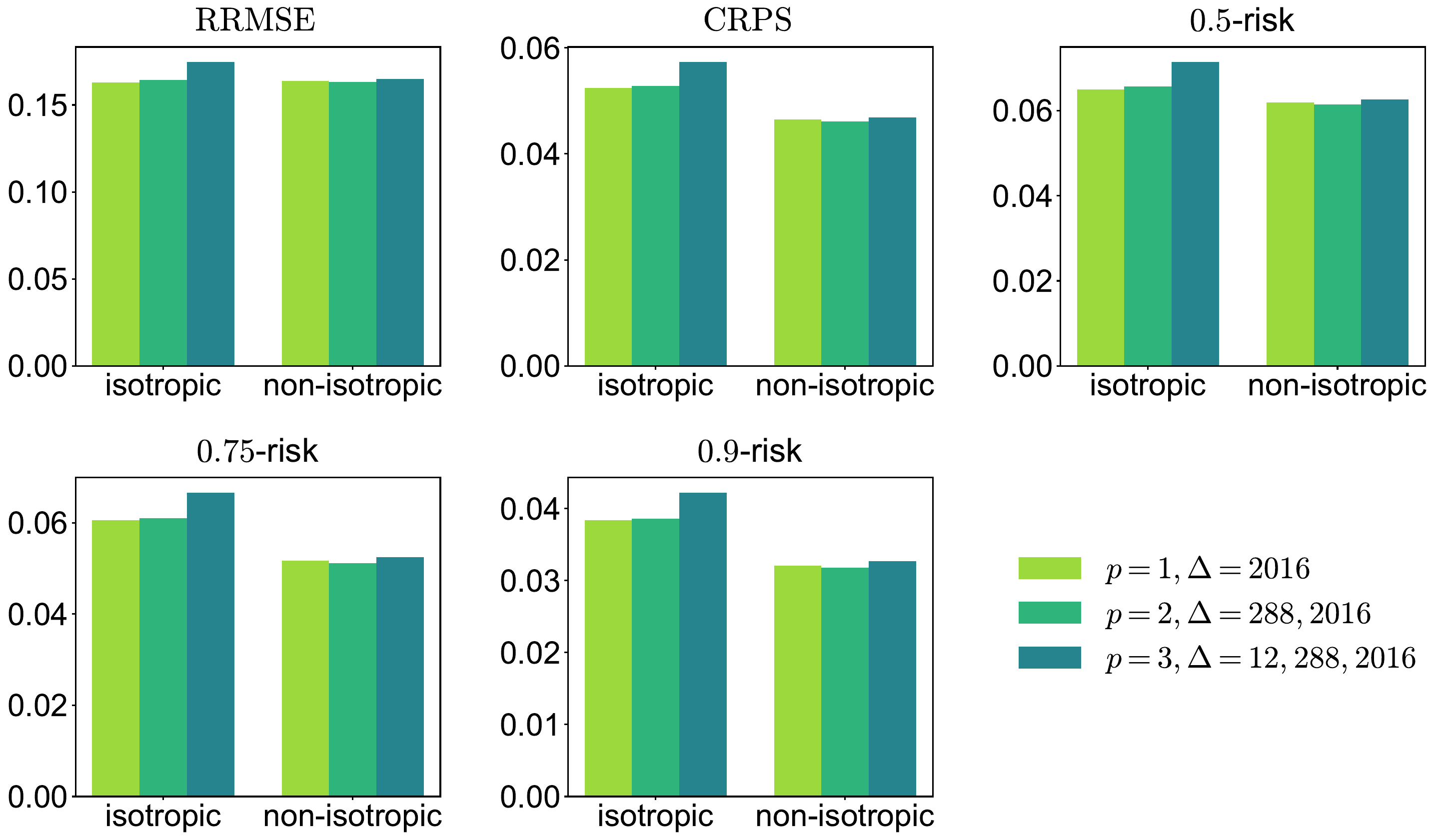}
  \caption{Ablation studies on the order of the error AR process using the Graph WaveNet model on the PEMS08 dataset.}
  \label{fig:abla_ar_order}
\end{figure}

To validate the first assumption, we test higher orders of the error AR process as shown in Figure~\ref{fig:abla_ar_order}. Using more lags in the error AR process does not necessarily improve performance and may degrade it with inappropriate lags (i.e., \(\Delta=12\) for the PEMS08 dataset). This is evident in Figure~\ref{fig:fig2_gwave_pm08_res}, where only the lag-2016 residuals are found to be correlated with the current residuals. Interestingly, the proposed non-isotropic loss demonstrates greater robustness to misspecification in the error AR process. As shown in Figure~\ref{fig:abla_ar_order}, when trained with non-isotropic loss, the inappropriate lag \(\Delta=12\) does not negatively influence the model as it does with isotropic loss. This is likely attributed to the greater flexibility that the non-isotropic loss provides during training.

\begin{figure}[htbp]
  \centering
  \includegraphics[width=0.7\textwidth, interpolate=false]{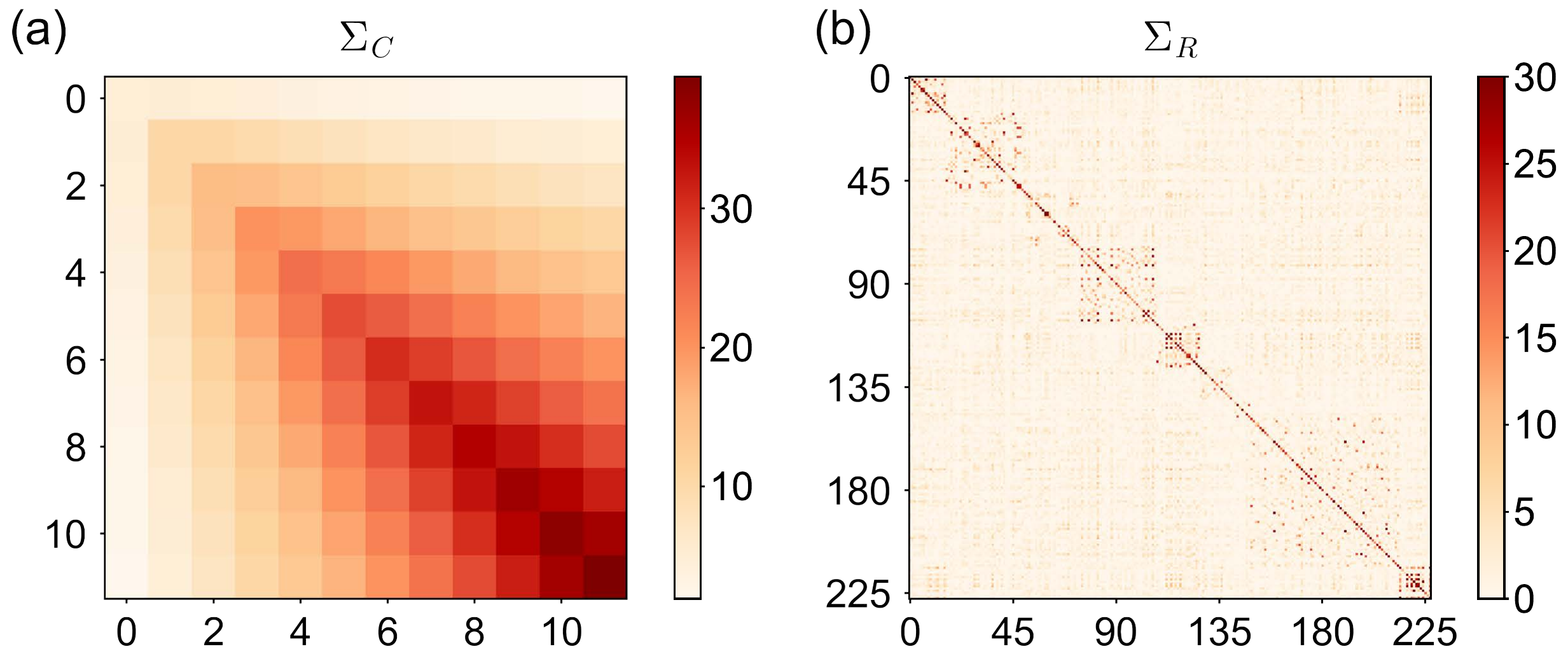}
  \caption{Empirical covariance matrices \(\boldsymbol{\Sigma}_R\) and \(\boldsymbol{\Sigma}_C\) calculated using the residuals of the Graph WaveNet model on the PEMSD7(M) dataset. Note that the value scales differ from those in Figure~\ref{fig:fig5_gwave_pm7m_cov} because the residuals are presented in the original data scales.}
  \label{fig:gwave_pm7m_empirical_cov}
\end{figure}

To validate the second assumption, we compare the learned covariance structure with the empirical covariance of the residuals. We collect all residuals \(\boldsymbol{E}_t\) in the testing set and calculate the row covariance and column covariance by:
\begin{equation}
\begin{aligned}
    &\boldsymbol{E}_{\text{row}} = \left[\boldsymbol{E}_1, \boldsymbol{E}_2, \dots, \boldsymbol{E}_T\right] \in \mathbb{R}^{N \times (T \times Q)}\\
    &\boldsymbol{\Sigma}_R=\frac{1}{TQ-1}\boldsymbol{E}_{\text{row}}\boldsymbol{E}_{\text{row}}^\top
\end{aligned}
\end{equation}
\begin{equation}
\begin{aligned}
    &\boldsymbol{E}_{\text{col}} = \left[\boldsymbol{E}_1^\top, \boldsymbol{E}_2^\top, \dots, \boldsymbol{E}_T^\top\right] \in \mathbb{R}^{Q \times (T \times N)}\\
    &\boldsymbol{\Sigma}_C=\frac{1}{TN-1}\boldsymbol{E}_{\text{col}}\boldsymbol{E}_{\text{col}}^\top
\end{aligned}
\end{equation}
where \(T\) is the number of observations in the testing set. The results are presented in Figure~\ref{fig:gwave_pm7m_empirical_cov}. We observe that the empirical covariance matrices exhibit a very similar structure to the learned covariance matrices shown in Figure~\ref{fig:fig5_gwave_pm7m_cov}. The Kronecker product with a diagonal structure simplifies log-likelihood evaluation in Eq.~\eqref{eq:nll_mnd} using the Sherman–Morrison–Woodbury identity (Eq.~\eqref{eqn:sigma_inv} and Eq.~\eqref{eqn:sigma_det}). Without this parameterization, computing the log-likelihood involves evaluating the inverse and the determinant of \(\boldsymbol{\Sigma}\) with size \(NQ \times NQ\), for which a na\"{i}ve implementation has a prohibitive time complexity of \(\mathcal{O}\left(N^3Q^3\right)\). This is computationally prohibitive in most cases.

\subsubsection{Visualization of Predictions}

\begin{figure}[htbp]
  \centering
  \includegraphics[width=0.65\textwidth, interpolate=false]{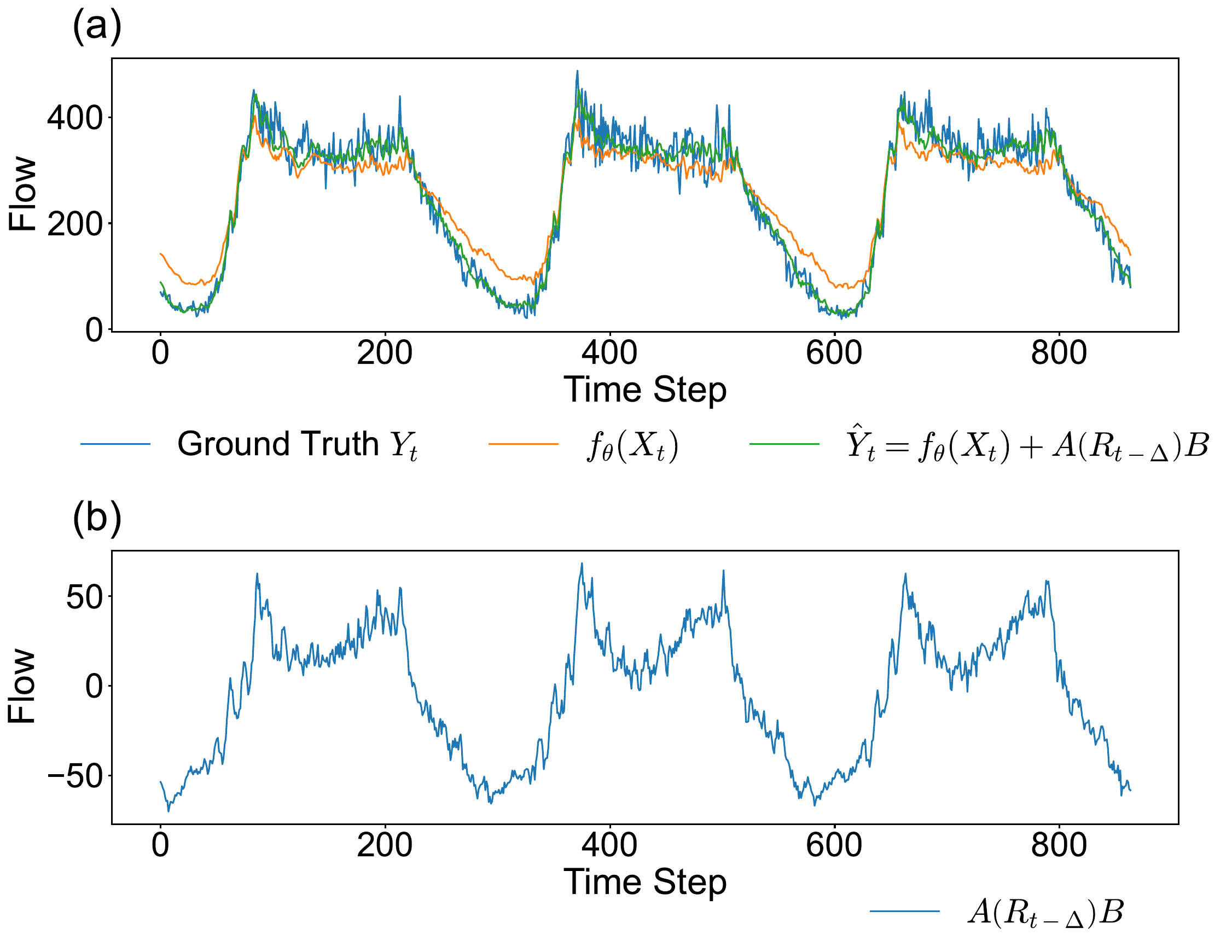}
  \caption{An example of the 6-step-ahead prediction results using the Graph WaveNet with the DR framework for the last three days of the PEMS08 dataset for one traffic sensor.}
  \label{fig:gwave_pm08_flow_compare}
\end{figure}

In this section, we evaluate our model by visualizing both deterministic and probabilistic forecast results. The deterministic forecast given by Eq.~\eqref{eqn:mean_pred} combines predictions from the DL model \(f_\theta\) and the error AR process \(\boldsymbol{A}\boldsymbol{R}_{t-\Delta}\boldsymbol{B}\). Figure~\ref{fig:gwave_pm08_flow_compare} illustrates the breakdown of the forecasts made by our method, highlighting the complementary effects of the two components. The error AR process primarily focuses on seasonal characteristics, such as traffic flow during congested regimes, while the DL model captures short-range, rapidly changing patterns.

The probabilistic forecast is given by the sampling process outlined in Eq.~\eqref{eqn:sampling}. The predicted mean \(\hat{\boldsymbol{Y}}_t\) serves as the location parameter in the predictive distribution for \(\boldsymbol{Y}_t\), while \(\Tilde{\boldsymbol{E}_t}\) represents a sample of the error term, capturing uncertainties in the forecasts by incorporating contemporaneous correlations learned in Section \ref{sec:cov}. Notably, the Kronecker structure within the covariance matrix explicitly characterizes both the covariance over the prediction horizon and the covariance over the spatial locations. We compare predictions from four randomly selected sensors at a fixed time using the original model and our DR framework.

\begin{figure}[htbp]
  \centering
  \includegraphics[width=0.75\textwidth, interpolate=false]{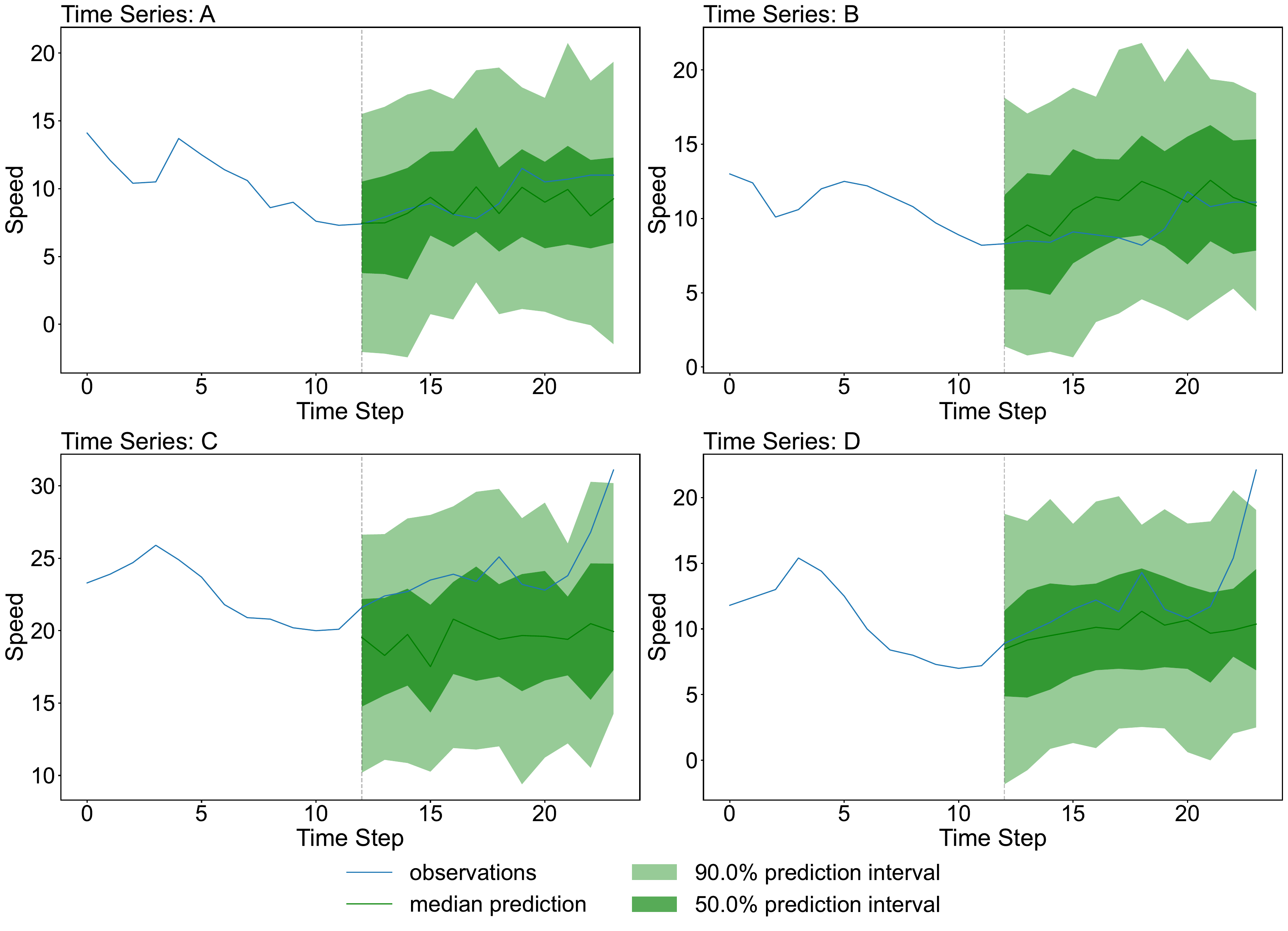}
  \caption{Probabilistic forecasting examples using Graph WaveNet on PEMSD7(M) without the integration of our proposed method. The dashed line splits the input time horizon ($P$) and the output time horizon ($Q$)}
  \label{fig:fig6_gwave_pm7m_prob}
\end{figure}

\begin{figure}[htbp]
  \centering

  \includegraphics[width=0.75\textwidth, interpolate=false]{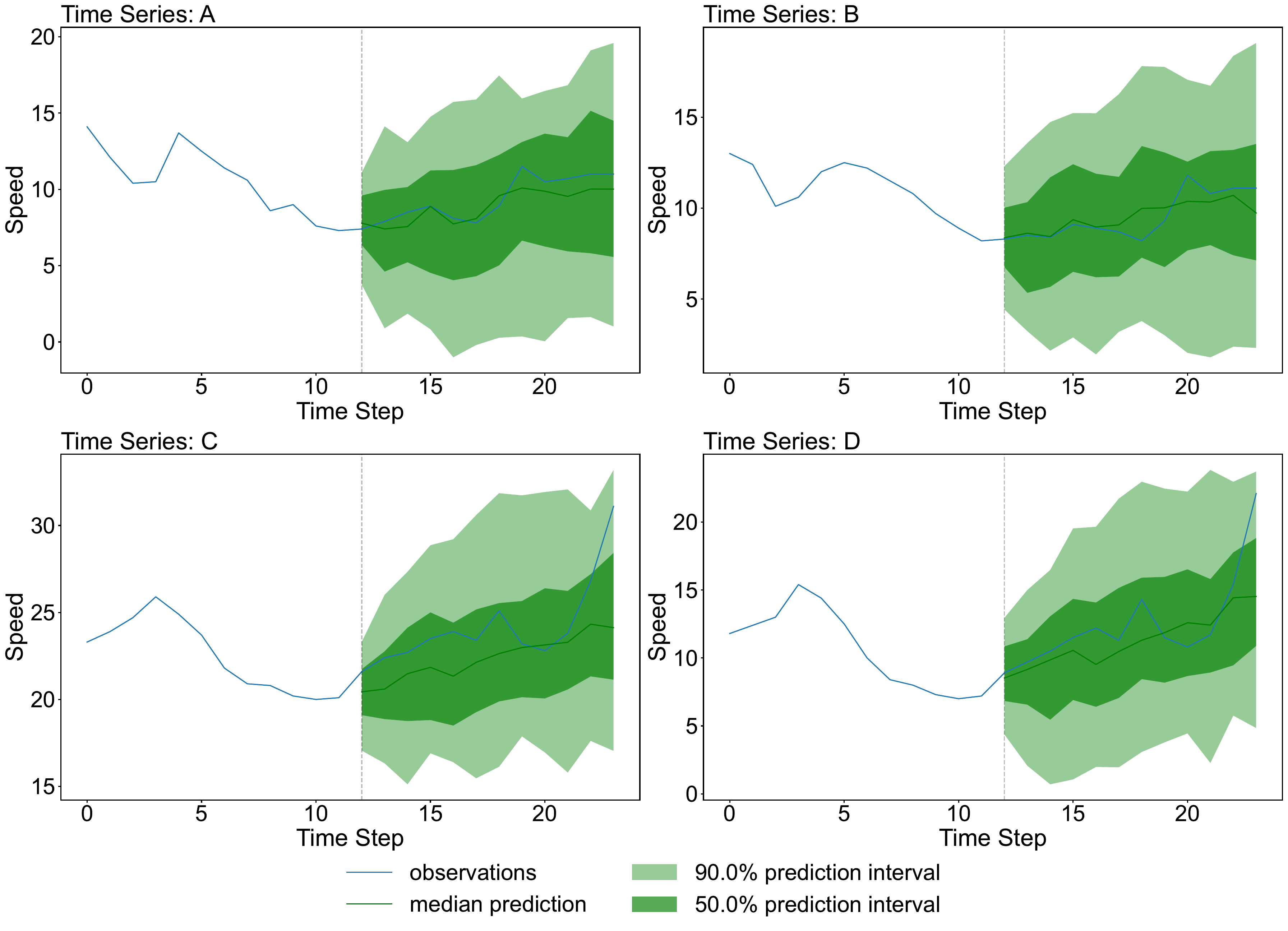}
  \caption{Probabilistic forecasting examples using Graph WaveNet on PEMSD7(M) with the integration of our proposed method. The dashed line splits the input time horizon ($P$) and the output time horizon ($Q$)}
  \label{fig:fig7_gwave_pm7m_prob_dr}
\end{figure}

Figure~\ref{fig:fig6_gwave_pm7m_prob} and Figure~\ref{fig:fig7_gwave_pm7m_prob_dr} offer a visual comparison of the probabilistic forecasting performance of Graph WaveNet on PEMSD7(M) with and without the incorporation of our method. In Figure~\ref{fig:fig6_gwave_pm7m_prob}, the original model assumes isotropic errors, with $\boldsymbol{\eta}_t$ following a Gaussian distribution $\mathcal{N}(\mathbf{0}, \sigma^2\mathbf{I}_{NQ})$. This results in a probabilistic forecasting  characterized by constant variance across the prediction horizon. However, such an approach proves unrealistic as it fails to account for the escalating uncertainty when making predictions into the future. This condition holds true for the spatial dimension as well, where all sensors exhibit the same variance, and no across-sensor correlations are considered.

In contrast, Figure~\ref{fig:fig7_gwave_pm7m_prob_dr} provides a visual representation of probabilistic forecasts when our DR method is implemented, incorporating both the calibration from error AR module and the learning of contemporaneous correlations. The impact of the error AR module is discernible in the relative position of the predictive distribution with respect to the observations. A notable instance is the comparison of time series C in Figure~\ref{fig:fig6_gwave_pm7m_prob} and Figure~\ref{fig:fig7_gwave_pm7m_prob_dr}, where the predictive distribution in Figure~\ref{fig:fig7_gwave_pm7m_prob_dr} exhibit a better alignment with the ground truth observations.

Furthermore, the explicit effects of modeling non-isotropic errors are evident in the widened uncertainty over the prediction horizon, as well as the diverse uncertainties observed for different time series at the same prediction step. Although the modeling of spatial correlations may not be directly observable, these factors are statistically accounted for during the sampling process. The noticeable variance disparity between the first and last prediction steps underscores the enhanced representation of uncertainty over time achieved through our proposed method.

\section{Discussion}\label{sec:discuss}
STF models, as they advance, can theoretically approximate independent errors. However, our framework highlights two alternative paths for STF models:

\textbf{Modeling Error Dynamics Separately.} The intentional separation between the base STF model and the error dynamics in the proposed DR framework explicitly divides the tasks into: 1) learning the primary spatiotemporal dependencies using the base model; and 2) capturing the error dynamics, including error correlations across time, space, and the prediction horizon, using the DR framework. This modular design acknowledges that, while an STF model may approximate independent errors, achieving this in practice remains challenging due to the complexity of spatiotemporal feature engineering and model design. Our framework provides a plug-and-play enhancement for any STF model, requiring no major architectural changes. A more detailed discussion is provided in Appendix \ref{apx:benefit1}.

\textbf{Advancing Probabilistic Forecasts.} Deterministic frameworks, constrained by isotropic assumptions, fail to capture two key phenomena in spatiotemporal traffic forecasting: 1) the growing uncertainty of predictions over the prediction horizon, represented by $\boldsymbol{\Sigma}_Q$ in our framework; and 2) the varying levels of variance across different spatial locations, represented by the diagonal elements of $\boldsymbol{\Sigma}_N$. In contrast, our framework serves as a model-agnostic probabilistic enhancement that can be seamlessly integrated into any STF model, regardless of its deterministic performance. A more detailed discussion is provided in Appendix \ref{apx:benefit2}.

To summarize, the conceptual contribution of our work lies in explicitly recognizing and efficiently modeling error correlations, rather than relying on the implicit assumption that a single STF model can fully capture them. This modular approach improves interpretability by revealing key error structures, such as increasing uncertainty over time and spatial error correlations, which remain obscured in end-to-end deep learning models.

\section{Conclusion}\label{sec:summary}
This paper presents a dynamic regression framework that enhances deep spatiotemporal models for probabilistic traffic forecasting by relaxing assumptions of time-independent and isotropic errors. We adjust the loss function to account for correlated errors, enabling seamless integration with existing models. Specifically, we represent the error series using a first-order matrix-variate autoregressive process with a non-isotropic error term. The new parameters introduced by our method can be jointly learned with the base forecasting model. Extensive experiments with state-of-the-art traffic forecasting models on real-world datasets validate the effectiveness of our DR framework. The acquired covariance matrix also facilitates probabilistic forecasting with enhanced uncertainty quantification for the multivariate Seq2Seq forecasting task. Additionally, the learned parameters offer clear physical and statistical interpretations.

This study has two primary limitations. First, learning the enriched error covariance structure increases training costs. For instance, applying our method to Graph WaveNet on PEMS08 increased training time by nearly 20\%. Second, scalability challenges may persist when applying our method to datasets with very large spatial dimensions, despite optimization efforts. Future work can focus on scaling our method for higher-dimensional datasets and incorporating time-varying error covariance matrices.

This work is a first step in addressing spatiotemporal error correlation in multivariate Seq2Seq deep learning models for traffic forecasting, contributing to the field. While developed for traffic forecasting, our method can be adapted to broader spatiotemporal forecasting tasks, such as daily weather and climatology prediction.

\section*{Acknowledgements}

The research is supported by the Natural Sciences and Engineering Research Council (NSERC) of Canada (Discovery Grant RGPIN 2019-05950). Vincent Zhihao Zheng also acknowledges the support received from the FRQNT B2X Doctoral Scholarship Program. The authors are grateful to the Associate Editor and three anonymous referees for valuable comments on an earlier version of the paper.

\bibliography{refs}

\pagebreak

\appendix
\section{Benefits of Separately Modeling the Error Dynamics}\label{apx:benefit1}
This section discusses the benefits of the intentional separation between the base GNN model and the error dynamics in the proposed DR framework. This separation explicitly divides the tasks into:
\begin{enumerate}[nosep, noitemsep]
    \item Learning the primary spatiotemporal dependencies (with the base GNN model).
    \item Capturing the error dynamics, including error correlations across time, space, and prediction horizon (with the DR framework).
\end{enumerate}
This modular design recognizes that while GNN models may approximate independent errors, achieving this in practice is challenging due to the complexity of spatiotemporal feature engineering and model design. Our framework serves as a plug-and-play enhancement for any GNN model, improving performance without extensive redesign. We provide an example in Table~\ref{tab:efficiency} using a SOTA GNN model, D\textsuperscript{2}STGNN \citep{shao2022decoupled}, as shown in the LargeST benchmark \citep{liu2024largest}. We observe that D\textsuperscript{2}STGNN, a more recent model developed three years after Graph WaveNet, achieves only a marginal improvement in performance ($1.01\% \sim 1.69\%$ in $\operatorname{RRMSE}$, and $1.28\% \sim 2.57\%$ in $\operatorname{CRPS}$) despite a significant increase (around $27.7\%$) in parameter size and the introduction of a new model design. In comparison, when enhanced with our framework, Graph WaveNet achieves a comparable improvement in $\operatorname{RRMSE}$, the deterministic score, with a smaller parameter increase ($16.8\% \sim 18.8\%$) and without the need for rigorous model redesign. Crucially, our framework significantly improves probabilistic forecasts, as reflected in the $\operatorname{CRPS}$.

\begin{table*}[htbp]
\caption{Efficiency analysis of the DR framework. The DR framework can enhance an older model to achieve the performance of a SOTA model in $\operatorname{RRMSE}$ while significantly improving probabilistic performance in $\operatorname{CRPS}$. The relative changes are calculated based on Graph WaveNet.}\label{tab:efficiency}
\centering
\begin{tabular}{ccccc}
\toprule
Data & Model & Num. Params. & $\operatorname{RRMSE}$ & $\operatorname{CRPS}$  \\
\midrule
\multirow{3}{*}{\rotatebox[origin=c]{90}{PEMSD7}}
& Graph WaveNet (2019)          & 309,757                & 0.3676                       & 0.0427             \\
\cmidrule(lr){3-5}
& D\textsuperscript{2}STGNN (2022)                & 395,709 ($\uparrow$27.7\%)      & 0.3639 ($\downarrow$1.01\%)             & 0.0416 ($\downarrow$2.57\%)   \\
\cmidrule(lr){3-5}
& Graph WaveNet + DR            & 361,885 ($\uparrow$16.8\%)      & 0.3622 ($\downarrow$1.46\%)             & 0.0389 ($\downarrow$8.89\%)   \\
\midrule
\multirow{3}{*}{\rotatebox[origin=c]{90}{PEMS08}}
& Graph WaveNet (2019)          & 308,597                & 0.1707                       & 0.0544             \\
\cmidrule(lr){3-5}
& D\textsuperscript{2}STGNN (2022)                & 394,317 ($\uparrow$27.7\%)      & 0.1678 ($\downarrow$1.69\%)             & 0.0537 ($\downarrow$1.28\%)   \\
\cmidrule(lr){3-5}
& Graph WaveNet + DR            & 366,685 ($\uparrow$18.8\%)      & 0.1639 ($\downarrow$3.98\%)             & 0.0465 ($\downarrow$14.5\%)    \\
\bottomrule
\end{tabular}
\end{table*}

\section{Limitations of the Deterministic Framework}\label{apx:benefit2}

The common deterministic framework, constrained by its isotropic assumption, fails to capture two key phenomena in spatiotemporal traffic forecasting: increasing uncertainty over the prediction horizon (\(\boldsymbol{\Sigma}_Q\)) and spatially varying variance (\(\boldsymbol{\Sigma}_N\)). In contrast, our framework can serve as a model-agnostic probabilistic enhancement to any SOTA GNN, irrespective of its deterministic performance. As shown in Table~\ref{tab:d2stgnn_benchmarks}, our method can still achieve significant improvements in probabilistic scores when applied to the SOTA D\textsuperscript{2}STGNN.

\begin{table*}[htbp]
\caption{Performance comparison of D\textsuperscript{2}STGNN with and without the DR framework. The best results are highlighted in \textbf{bold}.}\label{tab:d2stgnn_benchmarks}
\centering
\begin{tabular}{ccccccc}
\toprule
Data & Model & $\operatorname{RRMSE}$ & $\operatorname{CRPS}$ & $0.5$-risk & $0.75$-risk & $0.9$-risk\\
\midrule
\multirow{3}{*}{\rotatebox[origin=c]{90}{PEMSD7}}
 & D\textsuperscript{2}STGNN            & 0.3639      & 0.0416          & 0.0489          & 0.0452          & 0.0305          \\
 & \quad\quad + DR      & \textbf{0.3592}         & \textbf{0.0391} & \textbf{0.0481} & \textbf{0.0407} & \textbf{0.0271} \\
 \cmidrule(lr){2-7}
 & \quad\quad rel. impr.  & 1.29\%         & 6.01\%          & 1.64\%          & 9.96\%          & 11.15\% \\
\midrule
\multirow{3}{*}{\rotatebox[origin=c]{90}{PEMS08}}
 & D\textsuperscript{2}STGNN            & 0.1678          & 0.0537          & 0.0683          & 0.0615          & 0.0388          \\
 & \quad\quad + DR      & \textbf{0.1665}         & \textbf{0.0481} & \textbf{0.0645} & \textbf{0.0545} & \textbf{0.0331} \\
 \cmidrule(lr){2-7}
 & \quad\quad rel. impr.   & 0.77\%         & 10.43\%         & 5.56\%          & 11.38\%         & 14.69\% \\
\bottomrule
\end{tabular}
\end{table*}

To summarize, the conceptual contribution of our work lies in explicitly acknowledging and parsimoniously modeling error correlations, rather than relying on the implicit assumption that a single deterministic model (GNN alone) can fully capture them. This modular approach significantly enhances interpretability, as the DR framework provides valuable insights into the error structure (e.g., growing uncertainty over time and spatial error correlations) that would otherwise remain hidden in a purely end-to-end GNN model. Moreover, our framework remains consistent with deterministic assumptions, as its parameters adapt to different models and datasets. It reduces to the original model when \(\boldsymbol{A} = \mathbf{0}\) and \(\boldsymbol{\Sigma}_Q = \boldsymbol{\Sigma}_N = \mathbf{0}\).

\section{Consistency with Traffic Flow Theory}\label{apx:consistency_ft}

\begin{figure}[htbp]
  \centering
  \includegraphics[width=0.99\textwidth, interpolate=false]{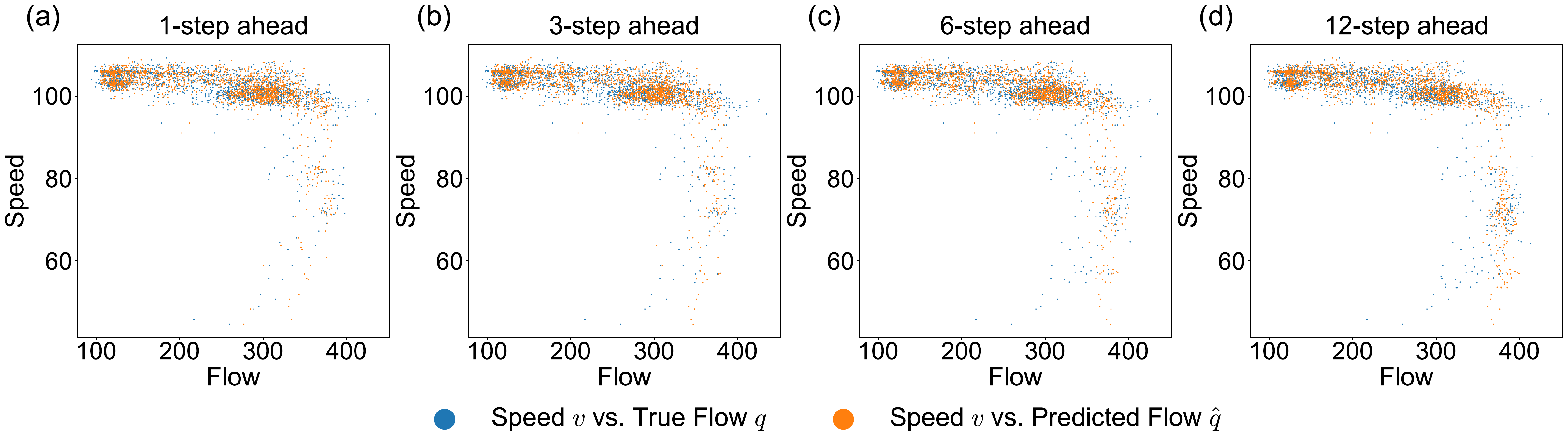}
  \caption{Comparison of the speed-flow relationship using ground truth data and predicted values during workday non-rush hours. Results are obtained using the Graph WaveNet model on the PEMS08 dataset.}
  \label{fig:gwave_pm08_workdaynorush}
\end{figure}

\begin{figure}[htbp]
  \centering
  \includegraphics[width=0.99\textwidth, interpolate=false]{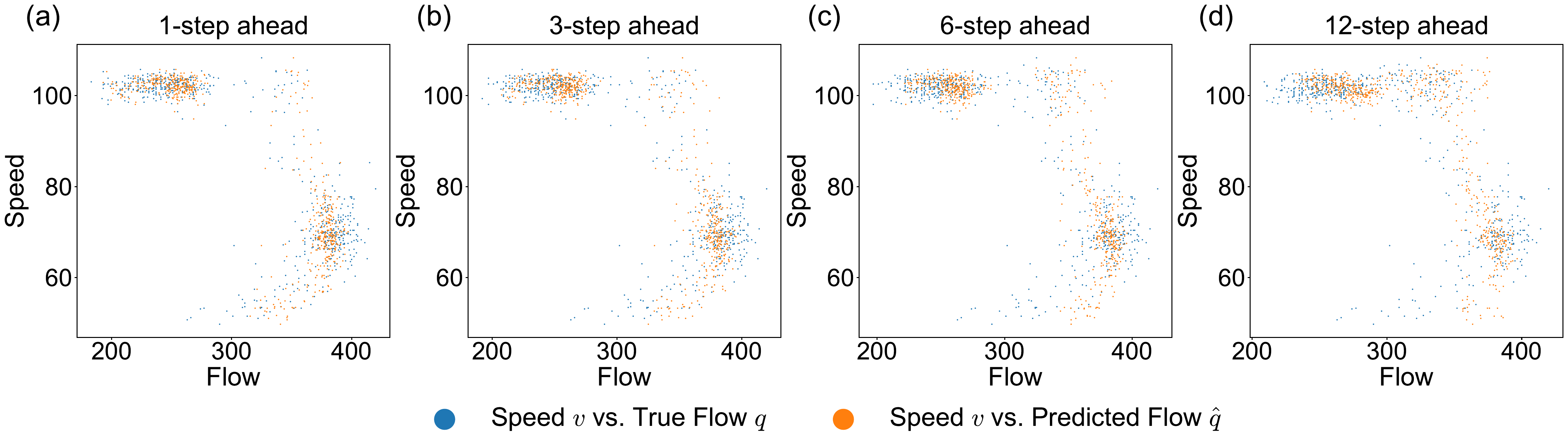}
  \caption{Comparison of the speed-flow relationship using ground truth data and predicted values during workday rush hours. Results are obtained using the Graph WaveNet model on the PEMS08 dataset.}
  \label{fig:gwave_pm08_workdayrush}
\end{figure}

\begin{figure}[htbp]
  \centering
  \includegraphics[width=0.99\textwidth, interpolate=false]{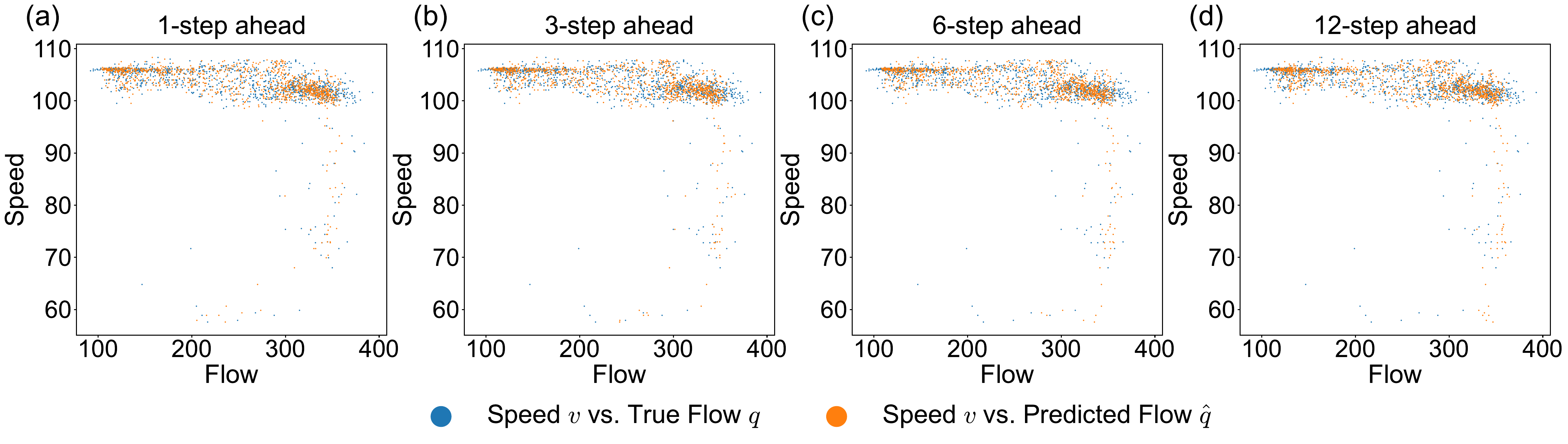}
  \caption{Comparison of the speed-flow relationship using ground truth data and predicted values during holiday. Results are obtained using the Graph WaveNet model on the PEMS08 dataset.}
  \label{fig:gwave_pm08_holiday}
\end{figure}

To validate consistency with traffic flow theory, we compare the speed-flow relationship from PEMS08 ground truth data with predictions from Graph WaveNet trained with our DR framework. We analyze three traffic scenarios: workday non-rush hours (Figure~\ref{fig:gwave_pm08_workdaynorush}), workday rush hours (Figure~\ref{fig:gwave_pm08_workdayrush}), and holidays (Figure~\ref{fig:gwave_pm08_holiday}). Results indicate strong alignment between predicted and ground truth speed-flow relationships across scenarios.

\section{Additional Plots}\label{apx:add_plots}

\begin{figure}[htbp]
  \centering
  \includegraphics[width=0.75\textwidth, interpolate=false]{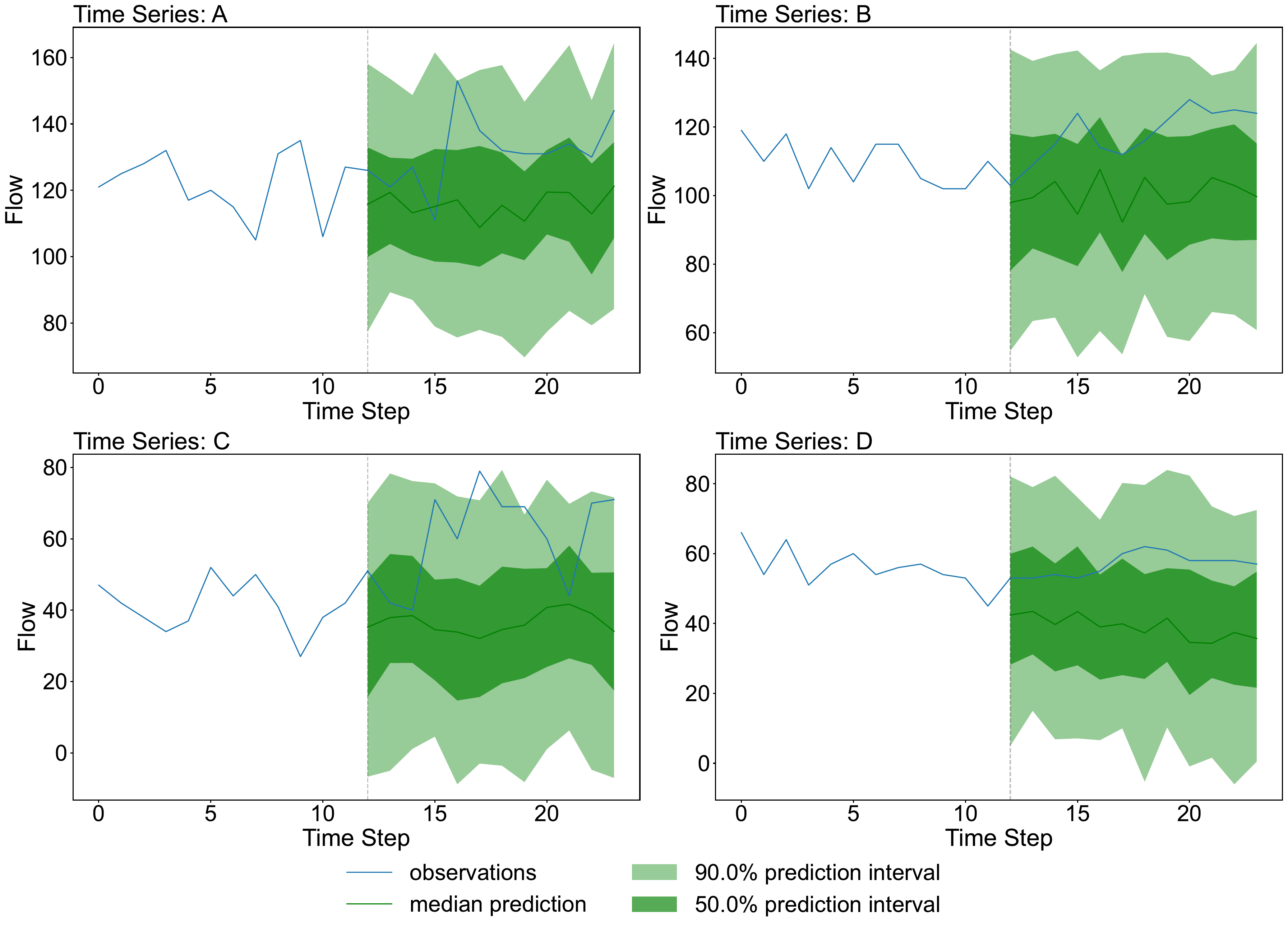}
  \caption{Probabilistic forecasting examples using Graph WaveNet on PEMS08 without the integration of our proposed method. The dashed line splits the input time horizon ($P$) and the output time horizon ($Q$)}
  \label{fig:fig8_gwave_pm08_prob}
\end{figure}

\begin{figure}[htbp]
  \centering

  \includegraphics[width=0.75\textwidth, interpolate=false]{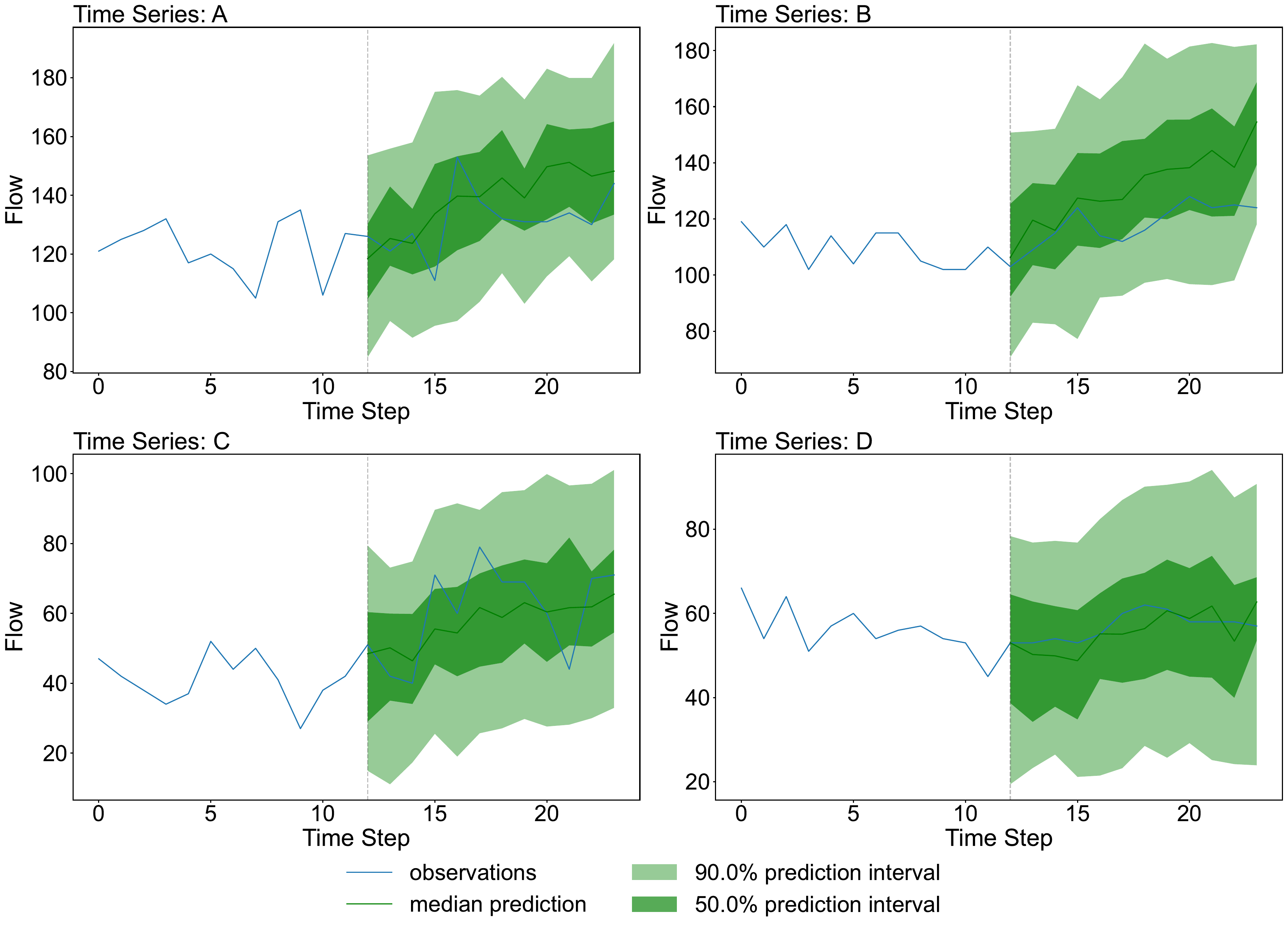}
  \caption{Probabilistic forecasting examples using Graph WaveNet on PEMS08 with the integration of our proposed method. The dashed line splits the input time horizon ($P$) and the output time horizon ($Q$)}
  \label{fig:fig9_gwave_pm08_prob_dr}
\end{figure}

We present an additional comparison for Graph WaveNet on PEMS08 in Figure~\ref{fig:fig8_gwave_pm08_prob} and Figure~\ref{fig:fig9_gwave_pm08_prob_dr}. Notably, our model demonstrates improved alignment between the predictive distribution and observations. Specifically, in Figure~\ref{fig:fig9_gwave_pm08_prob_dr}, the 50\% prediction intervals consistently encompass the ground truth observations across various forecasting steps, unlike Figure~\ref{fig:fig8_gwave_pm08_prob}. The variance increase is less pronounced in flow data than in speed data, as evident in the comparison between Figure~\ref{fig:fig7_gwave_pm7m_prob_dr} and Figure~\ref{fig:fig9_gwave_pm08_prob_dr}. This observation may suggest that predicting future values in the flow dataset involves less uncertainty than in the speed dataset, possibly due to the stronger seasonality effect in the flow data providing more information for longer-term forecasts. This also aligns with the indication of the optimal $\Delta=2016$ for PEMS08 in our error AR module.

\end{document}